\documentclass{article} %

\newif\ifsubmission
\newif\ifanon
\newif\ifshowtbd
\submissionfalse
\anonfalse
\showtbdfalse

\PassOptionsToPackage{table}{xcolor} %
\usepackage{iclr2027_conference,times}

\usepackage[utf8]{inputenc} %
\usepackage[T1]{fontenc}    %

\usepackage{float}          %
\usepackage[breaklinks=true,hidelinks]{hyperref} %
\usepackage{url}            %
\usepackage{booktabs}       %
\usepackage{array}          %
\usepackage{pifont}         %
\usepackage{multirow}       %
\usepackage{amsmath}        %
\usepackage{amssymb}        %
\usepackage{amsfonts}       %
\usepackage{amsthm}         %
\usepackage{graphicx}       %
\usepackage{nicefrac}       %
\usepackage{microtype}      %
\usepackage{xcolor}         %
\usepackage[most]{tcolorbox} %
\usepackage{enumitem}       %
\usepackage{wrapfig}        %
\usepackage{needspace}      %
\usepackage{placeins}       %
\usepackage{capt-of}        %
\usepackage{comment}        %

\floatstyle{ruled}
\newfloat{algorithm}{t}{loa}
\floatname{algorithm}{Algorithm}

\DeclareMathAlphabet\mathbfcal{OMS}{cmsy}{b}{n}

\newcommand{\mat}[1]{\mathbf{#1}}

\definecolor{crefold}{HTML}{3F5FA8}
\definecolor{ctask}{HTML}{2E7D32}
\definecolor{cwrap}{gray}{0.62}
\lstdefinestyle{refoldtext}{basicstyle=\ttfamily\small, breaklines=true, breakatwhitespace=true,
  upquote=true, aboveskip=0pt, belowskip=0pt, breakindent=0pt, columns=fullflexible, keepspaces=true,
  extendedchars=true,
  moredelim={[s][\color{crefold}\bfseries]{[context}{]}},
  moredelim={[s][\color{crefold}]{`}{`}},
  moredelim={[s][\color{black!60}\bfseries]{[step}{]}},
  moredelim={[s][\color{black!45}]{[...}{]}},
  moredelim={[s][\color{black!50}]{<returncode}{>}}, moredelim={[s][\color{black!50}]{</returncode}{>}},
  moredelim={[s][\color{black!50}]{<output}{>}}, moredelim={[s][\color{black!50}]{</output}{>}},
  literate={—}{{-{}-}}2 {·}{{\textperiodcentered}}1 {→}{{$\rightarrow$}}1
    {PROGRESS:}{{{\bfseries PROGRESS:}}}9 {STALE:}{{{\bfseries STALE:}}}6}
\lstdefinestyle{refoldcode}{style=refoldtext, breakindent=1em,
  postbreak=\mbox{\textcolor{cwrap}{$\hookrightarrow$}\space}}
\tcbset{refold listing/.style={listing only, breakable, enhanced, arc=1.2mm, boxrule=0.8pt,
    left=5pt, right=5pt, top=4pt, bottom=4pt, toptitle=1.6pt, bottomtitle=1.6pt,
    fonttitle=\bfseries\small, coltitle=white, listing options={style=refoldtext}},
  refold code/.style={refold listing, listing options={style=refoldcode}}}
\newtcblisting{promptbox}[1]{refold listing, title={#1},
  colframe=black!85, colbacktitle=black!85, colback=black!3}
\newtcblisting{notebox}[1]{refold code, title={#1},
  colframe=crefold, colbacktitle=crefold, colback=crefold!5}
\newtcblisting{outputbox}[1]{refold code, title={#1},
  colframe=cbase, colbacktitle=cbase, colback=black!2}
\newtcblisting{taskbox}[1]{refold code, title={#1},
  colframe=ctask, colbacktitle=ctask, colback=ctask!5}
\definecolor{cagent}{HTML}{B45309}
\newtcblisting{replybox}[1]{refold code, title={#1},
  colframe=cagent, colbacktitle=cagent, colback=cagent!5}

\definecolor{cgood}{HTML}{15803D}
\definecolor{cbad}{HTML}{B02418}
\definecolor{crun}{HTML}{C05621}
\newcommand{\gd}[1]{{\color{cgood}\scriptsize #1}}
\newcommand{\bd}[1]{{\color{cbad}\scriptsize #1}}
\newcommand{\nd}[1]{{\color{black!45}\scriptsize #1}}
\newcolumntype{P}[1]{>{\hskip#1\relax}r>{\hskip\dimexpr2.5pt-2\tabcolsep\relax}l<{\hskip#1\relax}}

\theoremstyle{definition}
\newtheorem{observation}{Observation}
\tcolorboxenvironment{observation}{%
  enhanced, breakable, colback=black!2, colframe=black!55,
  boxrule=0.5pt, arc=1.5pt, outer arc=1.5pt,
  left=7pt, right=7pt, top=5pt, bottom=5pt,
  before skip=8pt, after skip=4pt}

\definecolor{cbase}{HTML}{585858}
\definecolor{cclip}{HTML}{9E2A2B}
\definecolor{cmask}{HTML}{3F5FA8}
\definecolor{cprune}{HTML}{A672CB}
\definecolor{crecall}{HTML}{8E710F}
\definecolor{ccompact}{HTML}{3AA070}
\newcommand{\strat}[1]{\textcolor{c#1}{\texttt{#1}}}

\ifsubmission
  \includecomment{subonly}\excludecomment{arxivonly}
  \newcommand{\sub}[1]{#1}\newcommand{\arx}[1]{}
\else
  \excludecomment{subonly}\includecomment{arxivonly}
  \newcommand{\sub}[1]{}\newcommand{\arx}[1]{#1}
\fi

\ifshowtbd
  \newcommand{\tbd}[1]{\textcolor{red}{[\textsc{tbd}:\,#1]}}
  \newcommand{\running}[1]{\textcolor{crun}{[\textsc{re-running}:\,#1]}}
  
\else
  \newcommand{\tbd}[1]{}
  \newcommand{\running}[1]{}
  
\fi

\ifanon\else\iclrfinalcopy\fi

\title{ReFold: Training-Free Reversible Inter-Turn Context Folding for Long-Horizon Agents}

\ifanon
  \author{Anonymous authors \\ Paper under double-blind review}
\else
  \author{Yupeng Su$^{1}$ \quad Jiayi Tian$^{1}$ \quad Zheng Zhang$^{1}$ \quad Souvik Kundu$^{2}$ \\
    $^{1}$University of California, Santa Barbara \quad $^{2}$Intel}
\fi

\AtBeginDocument{%
  \setlength{\textfloatsep}{\baselineskip}%
  \setlength{\floatsep}{\baselineskip}%
  \setlength{\intextsep}{\baselineskip}%
  \setlength{\abovecaptionskip}{5pt}%
  \setlength{\belowcaptionskip}{0pt}%
  \raggedbottom}

\begin{document}

\maketitle
\lhead{Preprint}  %

\begin{abstract}
Long-horizon LLM agents act on an append-only interaction history that is re-sent to the model at every step, so the context and its cost grow with steps until the sessions exceed the context window.
Existing methods manage the context through context requirement prediction, relying on additional model calls, heuristic rules, or trained policies.
However, these predictive approaches introduce runtime overhead, invalidate prefix caches, and permanently discard content with no guarantee of recovery.
To overcome these limitations, we introduce \textbf{ReFold}: a \textit{training-free} rendering layer that preserves the underlying interaction history while compressing only the model's rendered context. It removes two kinds of inter-turn redundancy without an auxiliary predictor: content an earlier turn already displayed, replaced by a stub, and turns the agent itself reports finished, folded into a one-line note. Both operators use chunked rendering, rewriting the cached prefix once every few steps rather than at every step. Every removal is strictly reversible, a wrong removal costs one restore from the history rather than permanent content loss.
Because it operates at the rendering layer, ReFold is plug-and-play across standard ReAct-style harnesses. Evaluations across five long-horizon benchmarks and two frontier LLMs demonstrate that ReFold reduces token consumption by up to $2.5\times$ and halves the KV-cache memory per session without degrading task success rates. Under capped context budgets, it avoids up to $92\%$ of forced compactions. Under concurrent serving workloads, it reduces request queuing delays by up to $100\%$, accelerating inference by up to $1.7\times$, while cutting inference costs by $3.4\times$.
\end{abstract}

\begin{figure}[!h]
  \centering
  \includegraphics[width=\textwidth]{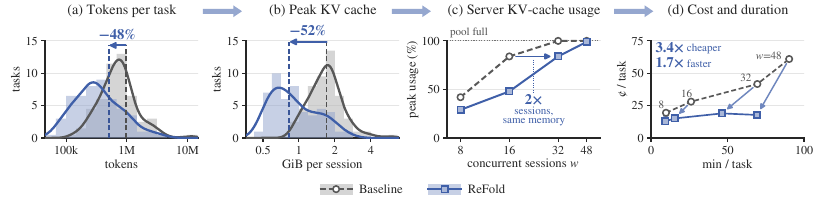}
  \caption{ReFold's savings carry from one session to the serving system. With \mbox{Qwen3.6-27B} on SWE-bench Verified, it cuts (a) tokens and (b) peak KV cache per session, so a vLLM server (c) fits $2\times$ the sessions in the same memory and (d) runs tasks up to $3.4\times$ cheaper and $1.7\times$ faster.}
  \label{fig:teaser}
\end{figure}

\section{Introduction}
\label{sec:intro}

Large language model (LLM) agents solve complex tasks by interleaving reasoning with actions and tool calls in an external environment \citep{yao2023react, schick2023toolformer, shinn2023reflexion}, and this framework has been widely adopted to push the boundaries of automated software engineering.
Given a natural-language issue and repository access \citep{jimenez2024swebench}, these systems localize faults, edit source files, and validate their own patches
\citep{yang2024sweagent, wang2024openhands, xia2024agentless}. 
Crucially, these workflows are inherently \emph{long-horizon}: resolving real-world repository issues requires continuous loops of reading, searching, testing, and editing, resulting in execution trajectories that span dozens and up to hundreds of steps \citep{merrill2026terminalbench}.

These agents operate over an append-only interaction history, resending the entire transcript of prior tool calls and environmental observations to the model at every step.
This history is dominated by tool outputs, such as file reads, search results, and test logs, which causes both context length and inference costs to grow with each successive step.
Furthermore, long contexts degrade model accuracy well before the window limit \citep{liu2024lost, hsieh2024ruler}. 
This pressure is directly evident at production scale: an analysis of GitHub Copilot over $95$ trillion tokens reveals that while only $7.8\%$ of agent sessions trigger context compaction, these sessions account for $44.2\%$ of total token usage \citep{liu2026agentic}. 
\textbf{Although sessions that exceed their context window represent a clear minority, they dominate overall serving costs}.

Existing methods prune the context by predicting whether past content will be needed later, and broadly fall into \textit{four} categories:
\ding{192} \emph{positional eviction}, which keeps only the most recent observations and masks older ones behind placeholders \citep{lindenbauer2025complexitytrap}; \ding{193} \emph{content pruning}, which drops observation content judged irrelevant by heuristic rules, trained scorers, or auxiliary model calls \citep{wang2026swepruner, xiao2026agentdiet, ren2026taco}; \ding{194} \emph{external storage}, which offloads evicted history to an addressable store the agent must query \citep{ehrlich2026lcm, arc2026}; and \ding{195} \emph{summarization}, which replaces past interactions with model-written summaries, either through an extra model call at a static token threshold, as in Claude Code and Codex, or through summaries the agent learns to write \citep{sun2025contextfolding, ye2026agentfold, gao2026swemem, li2026compactionrl}.
However, these predictive strategies come at a steep systemic cost: they introduce substantial runtime overhead, break prefix-cache reuse, or permanently discard context with no guarantee of recovery.

\textbf{Our Contributions.} To bridge these gaps, we present \textbf{ReFold}, a training-free rendering layer that preserves the agent's append-only interaction history while compressing only the context the model sees. Instead of an auxiliary predictor, ReFold removes content the history shows to be repeated and folds turns the agent itself reports finished, while avoiding the additional systemic costs identified above. We summarize our contributions as follows:

\begin{itemize}[leftmargin=1.4em,itemsep=2pt,topsep=3pt,parsep=0pt]
  \item We systematically evaluate six context-management strategies within a unified agent harness, showing that pruning tool outputs alone cannot bound a session's footprint and that predictive eviction incurs runtime overhead, invalidates prefix caches, or permanently discards context.

  \item We compress the rendered context through two lightweight operators. 
  \textit{Deduplication} replaces redundant content that previously appeared in earlier turns with a reference stub. \textit{Folding} collapses turns that the agent has itself reported as finished into a one-line note, spanning both tool observations and the agent's own messages.

  \item We avoid the systemic costs of existing strategies through two design choices. Both operators use chunked rendering, rewriting the cached prefix once every few steps rather than at every step. Every removal is strictly reversible by a restore from the unmodified history.
\end{itemize}

Evaluations across five long-horizon benchmarks and two frontier LLMs show that ReFold reduces token consumption and halves the per-session KV-cache footprint without degrading task success, and achieves lower cost and faster completion under capped budgets or concurrent load (Figure~\ref{fig:teaser}).

\section{Motivational Study}
\label{sec:motivation}
In this section, we examine the systemic costs of existing context-management
strategies through a controlled comparison in a unified agent
harness. Following the four categories introduced in
Section~\ref{sec:intro}, we implement four strategies: \ding{192}~masking observations older than a recency window behind placeholders (\strat{mask}), \ding{193}~pruning content within observations using heuristic rules or trained scorers (\strat{prune}), \ding{194}~evicting content into an addressable external store (\strat{recall}), and \ding{195}~compacting execution history via summarization (\strat{compact}). In addition, agent harnesses provide two baselines: \ding{196}~clipping older observations to a fixed character cap (\strat{clip}) and \ding{197}~retaining the interaction history unpruned (\strat{base}). All six variants are evaluated on the same $50$ SWE-bench Verified tasks under a $16$k context budget, with configurable hyperparameters
chosen by resolve rate (Appendix~\ref{sec:appendix-motivation}).

\begin{figure}[t]
  \centering
  \includegraphics[width=\textwidth]{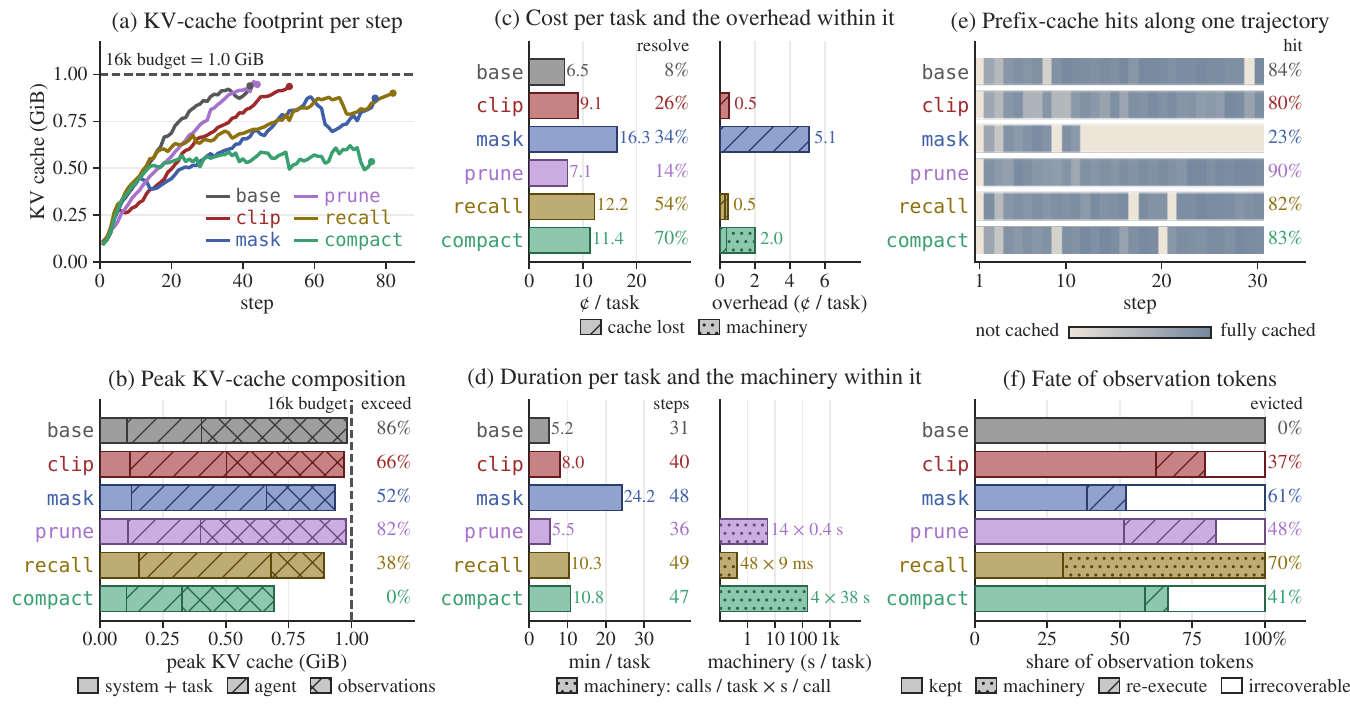}
\caption{The six strategies on the same $50$ SWE-bench Verified tasks at a $16$k budget: per-session KV-cache footprint and peak composition (a,~b), per-task cost and duration with resolve rate (c,~d), prefix-cache reuse along one trajectory (e) and the fate of evicted observation tokens (f).}
  \label{fig:sec2}
\end{figure}

\begin{observation}
Pruning tool outputs alone cannot guarantee that the context
remains within a fixed budget.
Enforcing a strict context bound requires controlling the
accumulation of both tool outputs and the agent's own commands
and reasoning traces in the rendered context.
\end{observation}

\noindent
The peak KV-cache memory required per session constrains how
many sessions a GPU can serve concurrently.
Figure~\ref{fig:sec2}a shows that the per-session footprint grows steadily with steps under all five non-compacting strategies.
In Figure~\ref{fig:sec2}b, these strategies reach peak footprints of approximately $0.89$--$0.98$\,GiB, close to the $1.0$\,GiB limit of the $16$k context budget in our setup.
Their context-budget exceed rates range from $38\%$ to $86\%$.
The component breakdown in Figure~\ref{fig:sec2}b explains why pruning tool outputs alone is insufficient.
Tool outputs account for approximately $59\%$ of the peak KV-cache footprint under \strat{base}. Under \strat{mask} and \strat{recall}, this share falls to approximately $24$--$29\%$, yet the total peak footprints remain close to the budget.
These methods leave the remaining $66$--$68\%$ of the budget untouched, including the fixed system prompt and task description as well as the agent's commands and reasoning traces, which accumulate with each step.
In contrast, \strat{compact}, which compresses both agent messages and tool outputs, reduces the peak KV-cache footprint to approximately $0.69$\,GiB with a $0\%$ context-budget exceed rate.
These empirical findings motivate us to extend compression beyond tool outputs to the agent's own commands and reasoning traces.

\begin{observation}
Auxiliary operations in context-management strategies introduce runtime overhead, adding inference cost and latency that offset part of the benefit from token savings.
\end{observation}

\noindent
Context-management strategies incur runtime overhead through auxiliary computation and disruptions to prefix-cache reuse.
Figure~\ref{fig:sec2}c and~d quantify these overheads in cost and execution time.
As Figure~\ref{fig:sec2}d shows, \strat{compact} makes a median of four summarization calls per task at approximately $38$\,s each, adding roughly $152$\,s to execution.
According to Figure~\ref{fig:sec2}c, these calls cost approximately $1.6$ cents per task, accounting for $14\%$ of its total $11.4$-cent cost.
Meanwhile, \strat{prune} invokes an auxiliary $0.6$B scorer $14$ times per task at approximately $0.4$ seconds per call, adding roughly $5$ seconds of latency that does not appear in the commercial token bill.
Even without auxiliary model calls, \strat{mask} increases inference cost by repeatedly modifying historical observations and invalidating cached prefixes: the lost cache alone adds $5.1$ cents per task, which accounts for $31\%$ of total cost.
The net benefit therefore depends on three factors: the tokens saved, the frequency and per-call overhead of auxiliary operations, and the prefix-cache reuse preserved during execution.
Learned strategies additionally require offline investment in training data, policy training, and adaptation to the target model and tool set.
Together, these costs motivate a training-free, data-free approach that reduces context without auxiliary model calls and adds minimal work to the execution loop.

\begin{observation}
Context eviction introduces two systemic costs: it invalidates reusable KV-cache prefixes and removes historical information without guaranteeing its recovery.
\end{observation}

\noindent
Modern serving frameworks reduce prefill computation by reusing cached prefixes from prior steps.
However, modifying historical context invalidates cached prefixes from the point of mutation onward, and the penalty depends on the position and frequency of these edits.
Figure~\ref{fig:sec2}e traces prefix-cache reuse along one trajectory. Under \strat{mask}, older observations are masked as they leave the recency window, repeatedly invalidating the cached prefix and reducing the cache-hit rate to $23\%$, compared with $84\%$ under \strat{base}.
\strat{compact} and \strat{recall} also interrupt prefix-cache reuse when they update historical context, although cache-hit rates recover between updates.
Beyond the cache penalty, eviction can permanently discard content that may remain relevant to subsequent steps.
Figure~\ref{fig:sec2}f classifies an evicted token as recoverable if the strategy's own mechanism can return it or re-executing a read of an unchanged file can reproduce it, and as irrecoverable otherwise.
\strat{clip}, \strat{mask}, \strat{prune}, and \strat{compact} have no such mechanism: they evict $37$--$61\%$ of the observation tokens, and $17$--$48\%$ of all observation tokens cannot be recovered, whereas \strat{recall} can return everything it evicts.
Current eviction thus buys shorter prompts at the cost of prefix-cache reuse and recoverability, which motivates limiting prefix changes while keeping every eviction recoverable.

\section{Method: ReFold}
\label{sec:method}

\begin{figure}[t]
  \centering
  \includegraphics[width=\linewidth]{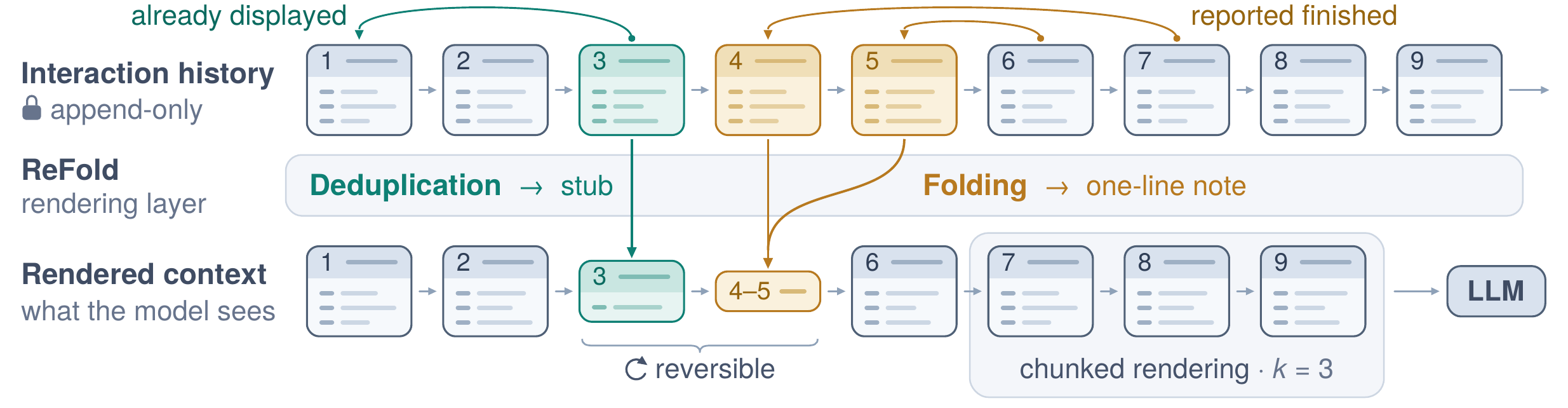}
  \caption{ReFold compresses the rendered context while preserving the append-only history. Turn~3 repeats turn~1 and becomes a stub, and turns~4--5, reported finished, collapse into a note. Turns~\mbox{7--9}, past the $k{=}3$ boundary, render in full. Stubs and notes can be restored from the history.}
  \label{fig:method}
\end{figure}

In this section, we present the design and implementation of ReFold (Figure~\ref{fig:method}).
It addresses the limitations mentioned above, with a rendering function that compresses the context sent to the model while preserving the append-only interaction history. 
Section~\ref{sec:method-setup} introduces the notation.
Section~\ref{sec:rendering} describes the two rendering operators, \textit{deduplication} and \textit{folding}.
Section~\ref{sec:method-window} explains their chunked rendering and reversibility, and Section~\ref{sec:method-impl} details ReFold's integration into the agent loop.

\subsection{Setup}
\label{sec:method-setup}

An LLM agent operates over sequential steps $t = 1, \dots, T$ on an append-only interaction history:
\begin{equation}
  \mat{h}_t = \mat{p}_{\text{sys}} \oplus \mat{p}_{\text{task}} \oplus
               \mat{z}_1 \oplus \cdots \oplus \mat{z}_t,
  \qquad
  \mat{z}_i = \mat{a}_i \oplus \mat{o}_i,
  \label{eq:history}
\end{equation}
where $\oplus$ denotes concatenation, $\mat{p}_{\text{sys}}$ is the system prompt, $\mat{p}_{\text{task}}$ is the task description,
and each turn $\mat{z}_i$ is formulated as a pair, coupling the agent's action $\mat{a}_i$ (encompassing its internal reasoning trace and tool invocations) with the resulting environment observation $\mat{o}_i$.
At step $t$, the LLM $\mathcal{M}$ receives the context $\mat{c}_{t-1}$ constructed from the previous interaction history $\mat{h}_{t-1}$ and generates $\mat{a}_t \sim \mathcal{M}(\cdot \mid \mat{c}_{t-1})$.
The environment executes the tool call in $\mat{a}_t$ and returns $\mat{o}_t$, and the completed turn $\mat{z}_t$ is appended to form $\mat{h}_t$.
By default, the input context is the direct concatenation of the entire history:
$\mat{c}_{t-1} = \mat{h}_{t-1}$.
Since each turn is resent at every subsequent step, assuming a constant average turn length, the cumulative context volume scales quadratically as $\sum_{t=1}^{T} \lvert \mat{c}_{t-1} \rvert = \Theta(T^2)$.
To tame this quadratic growth without losing past interactions, ReFold devises a rendering function $R$ that maps the history $\mat{h}_t$ to a compressed context $\mat{c}_t = R(\mat{h}_t)$, without modifying $\mat{h}_t$.

\subsection{Context Rendering}
\label{sec:rendering}
ReFold implements the rendering function $R$ via two targeted operators. \textit{Deduplication} removes repeated content from an observation $\mat{o}_i$, and \textit{folding} removes a whole turn $\mat{z}_i$, the action $\mat{a}_i$ together with its observation $\mat{o}_i$, once the agent has reported it as finished.

\paragraph{Deduplication.}
Deduplication eliminates redundant tool outputs across historical turns.
When identical command outputs or file ranges recur, ReFold retains the initial occurrence and replaces subsequent duplicates with lightweight reference stubs.
We partition the observation $\mat{o}_i$ into $m_i$ consecutive spans $s_{i,1},\dots,s_{i,m_i}$, where each span is classified as either novel or an exact duplicate of a prior span.
The deduplicated observation is
\begin{equation}
  d_i = \bigoplus_{\ell=1}^{m_i}
  \begin{cases}
    \mathrm{stub}_{i,\ell} & \text{if span } s_{i,\ell}
                                \text{ is a duplicate},\\
    s_{i,\ell}                & \text{otherwise}.
  \end{cases}
  \label{eq:dedup}
\end{equation}
Each stub records the omitted region and points to the turn containing the retained copy; for long spans, it also keeps the top-level class and function signatures. 
Search results and repeated command outputs are deduplicated in the same way by comparing their full text.
When the entire read is a duplicate, the stub has the following form:

\begin{tcolorbox}[
  width=\linewidth,
  colback=crefold!5,
  colframe=crefold,
  boxrule=0.4pt,
  arc=1mm,
  left=5pt,right=5pt,top=4pt,bottom=4pt,
  fontupper=\small\ttfamily
]
{}\textcolor{crefold}{\bfseries[context-manager]} <path>:<start>-<end> omitted:
identical to step <source step>. \textcolor{crefold}{\textasciigrave{}restore <step>\textasciigrave{}} shows it again.
\end{tcolorbox}

\paragraph{Folding.}
Folding removes turns that the agent has reported as finished.
To detect finished turns without costly auxiliary operations, ReFold prompts the agent to prepend two metadata fields to its reasoning and tool call $\tilde{\mat{a}}_t$: \textsc{progress}, a one-line description of what the current step does, and \textsc{stale}, a set of prior step indices whose outputs the agent no longer needs, written as text $\langle\textsc{stale}_t\rangle$:
\begin{equation}
  \mat{a}_t = \textsc{progress}_t \oplus \langle\textsc{stale}_t\rangle \oplus \tilde{\mat{a}}_t.
  \label{eq:prepend}
\end{equation}
To facilitate this indexing, the observation of each turn is explicitly tagged with its step identifier.
Both fields are generated as part of the regular response and require no additional model calls (Appendix~\ref{sec:appendix-method}).
ReFold only accepts \textsc{stale} indices within the preceding $k$ steps:
\begin{equation}
  S_t = S_{t-1} \cup \bigl(\textsc{stale}_t \cap \{t{-}k, \dots, t{-}1\}\bigr), \quad S_0 = \varnothing.
  \label{eq:finished}
\end{equation}
For each step $i \in S_t$, the full message and tool output are replaced by a short note:
\begin{equation}
  f_i = \textsc{progress}_i \oplus e_i,
  \label{eq:fold-note}
\end{equation}
where $e_i$ collects key details from the removed output, such as function signatures or search results.
If no such details are available, $e_i$ is empty.
Notes of adjacent turns share one line in the following format, with the detail block included only when $e_i$ is nonempty:

\begin{tcolorbox}[
  width=\linewidth,
  colback=crefold!5,
  colframe=crefold,
  boxrule=0.4pt,
  arc=1mm,
  left=5pt,right=5pt,top=4pt,bottom=4pt,
  fontupper=\small\ttfamily
]
{}\textcolor{crefold}{\bfseries[context-manager]} steps <i>-<j> folded: <PROGRESS i> $\cdot$ ... $\cdot$ <PROGRESS j>. \textcolor{crefold}{\textasciigrave{}restore <i>-<j>\textasciigrave{}} shows them again.\\
\hspace*{1em}<file>: <extracted function signatures>\\
\hspace*{1em}\$ <search command> -> <retained search hits>
\end{tcolorbox}

\subsection{Chunked Rendering and Reversibility}
\label{sec:method-window}

\paragraph{Chunked rendering.}
To preserve prefix-cache reuse, both operators share a single boundary $b_t$ that advances once every $k$ steps:
\begin{equation}
  b_t = \max\!\left(0,\,k\!\left\lfloor\frac{t-k}{k}\right\rfloor\right),
  \label{eq:boundary}
\end{equation}
where $t$ denotes the current execution step.
All turns $\mat{z}_i$ with $i > b_t$ are protected and rendered in full.
Earlier turns are divided into two categories: turns marked as finished are replaced by fold notes (Eq.~\ref{eq:fold-note}), whereas the rest are deduplicated (Eq.~\ref{eq:dedup}).
Together, these components define the rendering function introduced in Section~\ref{sec:method-setup}:
\begin{equation}
  \mat{c}_t = R(\mat{h}_t) = \mat{p}_{\text{sys}} \oplus \mat{p}_{\text{task}} \oplus \bigoplus_{i=1}^{t}
  \begin{cases}
    f_i                          & \text{if } i \le b_t \textit{ and } i \in S_t, \\
    \mat{a}_i \oplus \mat{o}_i   & \text{if } i > b_t, \\
    \mat{a}_i \oplus d_i     & \text{otherwise}.
  \end{cases}
  \label{eq:render}
\end{equation}
Between boundary advances, the rendered context remains strictly append-only.
Because historical turns remain invariant and new turns are merely appended to the sequence, the KV cache for the rendered prefix is preserved and fully reused across steps. We set the chunk size to $k{=}3$.

\paragraph{Reversibility.}
Because ReFold maintains an append-only history, anything it removes can be recovered through two distinct mechanisms.
First, since each stub and fold note explicitly identifies the step indices it replaces, the agent can issue the command \texttt{restore <step>}.
Rather than forwarding this command to the environment, the execution harness looks up the turn in the history and appends the original output as a new turn at the end of the context.
This guarantees that the existing prompt prefix remains invariant and cache-friendly.
Second, the agent can re-execute any original command whose prior output was compressed to observe the latest environment state. ReFold preserves these re-executed observations in full and exempts them from subsequent compression.

\subsection{Integration with Agentic Workflows}
\label{sec:method-impl}

Algorithm~\ref{alg:render} in Appendix~\ref{sec:appendix-method} integrates ReFold into the agent loop: \textsc{Observe} appends each completed turn $\mat{z}_t$ to the history $\mat{h}_t$, extracts $\textsc{progress}_t$ and $\textsc{stale}_t$ from $\mat{a}_t$, and updates the set $S_t$ of finished steps; \textsc{Render} constructs the compressed context $\mat{c}_t = R(\mat{h}_t)$ from $\mat{h}_t$ and $S_t$ before each model call.
Because ReFold operates entirely at the rendering layer, it leaves the agent’s tools, environment, and interaction protocol unchanged. Integration requires only a rendering hook and prompt instructions that elicit the two signals, making ReFold easy to integrate into existing ReAct-style harnesses.

\section{Experiments}
\label{sec:experiments}

\subsection{Setup}
\label{sec:setup}

\paragraph{Agent and models.}
We implement ReFold in mini-swe-agent, a ReAct-style agent that interleaves reasoning with shell commands to inspect files, modify code, and execute tests in the task environment.
We evaluate two backbone models: Qwen3.6-27B, a dense model, and Qwen3.6-35B-A3B, a mixture-of-experts (MoE) model.
Both models are served using vLLM with prefix caching enabled, each on two NVIDIA RTX 6000 Ada GPUs with tensor parallelism.
We use a temperature of $0$, a maximum of $16{,}384$ generated tokens per step, and a $250$-step limit per trajectory.
The full context window is $256$k tokens, of which $16$k are reserved for generation; the capped budgets in Section~\ref{sec:results-scarcity} are stated as usable context in the same way.
All configurations use the same tools and execution environment, and command outputs are truncated at $10{,}000$ characters (Appendix~\ref{sec:appendix-setup}).

\paragraph{Benchmarks.}
We evaluate on five benchmarks covering repository-level software engineering and terminal-based tasks, using $50$ tasks from each (Appendix~\ref{sec:appendix-benchmarks}).
SWE-bench Verified~\citep{jimenez2024swebench, openai2024swebenchverified}, SWE-bench Pro~\citep{deng2025swebenchpro}, and Multi-SWE-bench~\citep{zan2025multiswebench} evaluate agents on resolving issues in software repositories, requiring them to locate relevant code, implement changes, and validate their solutions.
Terminal-Bench 1.0 and 2.1~\citep{tbench2025core, merrill2026terminalbench, tbench2026v21} extend the evaluation to tasks performed through a terminal.
All tasks run in the official containers and are scored by the official evaluators.

\paragraph{Baselines.}
We evaluate ReFold against two context-management baselines.
\strat{base} retains the complete interaction history without compression.
\strat{compact} invokes an additional model call to summarize the history when context usage reaches $75\%$ of the available budget, and subsequent execution proceeds from the summary.
We compare \strat{compact} with and without ReFold under the same experimental settings.
At the full window, the compaction threshold is never reached; consequently, \strat{compact} and \strat{base} produce identical runs.

\paragraph{Evaluation metrics.}
We evaluate task performance, resource consumption, and execution efficiency.
Task performance is measured by \emph{resolve}, the number of tasks solved according to the official benchmark evaluators.
Resource consumption is measured by \emph{tokens}, the mean input and output tokens per task, \emph{cost}, the mean estimated inference cost per task including cached-input discounts, both counting auxiliary model calls, and \emph{KV cache}, the median peak KV-cache memory per session.
On the server, \emph{KV peak} is the maximum share of the KV-cache pool in use during a run.
Execution efficiency is measured by \emph{steps}, the median number of model calls per task, \emph{duration}, the median time from the first request to the final response of a task, \emph{compactions}, the mean number of summarize-and-restart events of \strat{compact} per task, and \emph{queue}, the mean scheduling delay per request.
Detailed metric definitions and measurement conventions are provided in Appendix~\ref{sec:appendix-metrics-def}.

\subsection{Main Results}
\label{sec:results-main}

\begin{table}[t]
  \caption{Comparison across five benchmarks and two models, with $50$ tasks per benchmark at the full context window and $w{=}8$. Parentheses indicate changes relative to \strat{base}.}
  \label{tab:main}
  \centering \small
  \setlength{\tabcolsep}{3.2pt}
  \renewcommand{\arraystretch}{1.14}
  \scalebox{0.880}{%
  \begin{tabular}{l*{5}{rP{0pt}}}
    \toprule
    & \multicolumn{3}{c}{Resolve} & \multicolumn{3}{c}{Steps} & \multicolumn{3}{c}{Tokens (k)}
      & \multicolumn{3}{c}{KV cache (GiB)} & \multicolumn{3}{c}{Cost (\textcent)} \\
    \cmidrule(lr){2-4}\cmidrule(lr){5-7}\cmidrule(lr){8-10}\cmidrule(lr){11-13}\cmidrule(lr){14-16}
    Benchmark
      & \multicolumn{1}{c}{\strat{base}} & \multicolumn{2}{c}{ReFold}
      & \multicolumn{1}{c}{\strat{base}} & \multicolumn{2}{c}{ReFold}
      & \multicolumn{1}{c}{\strat{base}} & \multicolumn{2}{c}{ReFold}
      & \multicolumn{1}{c}{\strat{base}} & \multicolumn{2}{c}{ReFold}
      & \multicolumn{1}{c}{\strat{base}} & \multicolumn{2}{c}{ReFold} \\
    \midrule
    \multicolumn{16}{l}{\emph{Qwen3.6-27B (dense)}} \\
    SWE-bench Verified & $38$ & $38$ & \nd{($\pm$0)} & $50$ & $41$ & \gd{($-$18\%)}
      & $965$ & $505$ & \gd{($-$48\%)} & $1.71$ & $0.83$ & \gd{($-$52\%)} & $19.5$ & $12.9$ & \gd{($-$34\%)} \\
    SWE-bench Pro & $5$ & $6$ & \nd{($+$1)} & $46$ & $50$ & \bd{($+$9\%)}
      & $1097$ & $899$ & \gd{($-$18\%)} & $1.96$ & $1.20$ & \gd{($-$39\%)} & $20.8$ & $20.5$ & \nd{($-$2\%)} \\
    Multi-SWE-bench & $25$ & $25$ & \nd{($\pm$0)} & $47$ & $44$ & \gd{($-$6\%)}
      & $1372$ & $956$ & \gd{($-$30\%)} & $1.55$ & $0.93$ & \gd{($-$40\%)} & $25.5$ & $22.3$ & \gd{($-$13\%)} \\
    Terminal-Bench 1.0 & $23$ & $23$ & \nd{($\pm$0)} & $20$ & $15$ & \gd{($-$25\%)}
      & $1279$ & $596$ & \gd{($-$53\%)} & $1.02$ & $0.58$ & \gd{($-$43\%)} & $27.4$ & $19.0$ & \gd{($-$31\%)} \\
    Terminal-Bench 2.1 & $23$ & $26$ & \nd{($+$3)} & $42$ & $29$ & \gd{($-$31\%)}
      & $3989$ & $1909$ & \gd{($-$52\%)} & $1.60$ & $0.68$ & \gd{($-$58\%)} & $75.7$ & $49.2$ & \gd{($-$35\%)} \\
    \midrule
    \multicolumn{16}{l}{\emph{Qwen3.6-35B-A3B (MoE)}} \\
    SWE-bench Verified & $36$ & $36$ & \nd{($\pm$0)} & $70$ & $61$ & \gd{($-$13\%)}
      & $2019$ & $1063$ & \gd{($-$47\%)} & $0.74$ & $0.42$ & \gd{($-$43\%)} & $13.1$ & $8.2$ & \gd{($-$37\%)} \\
    SWE-bench Pro & $6$ & $5$ & \nd{($-$1)} & $82$ & $66$ & \gd{($-$20\%)}
      & $3328$ & $1344$ & \gd{($-$60\%)} & $0.92$ & $0.41$ & \gd{($-$55\%)} & $19.8$ & $11.0$ & \gd{($-$45\%)} \\
    Multi-SWE-bench & $18$ & $21$ & \nd{($+$3)} & $89$ & $66$ & \gd{($-$26\%)}
      & $3487$ & $1755$ & \gd{($-$50\%)} & $0.84$ & $0.47$ & \gd{($-$44\%)} & $20.7$ & $12.5$ & \gd{($-$40\%)} \\
    Terminal-Bench 1.0 & $21$ & $23$ & \nd{($+$2)} & $25$ & $22$ & \gd{($-$12\%)}
      & $1871$ & $1094$ & \gd{($-$42\%)} & $0.38$ & $0.27$ & \gd{($-$30\%)} & $13.6$ & $11.1$ & \gd{($-$18\%)} \\
    Terminal-Bench 2.1 & $21$ & $20$ & \nd{($-$1)} & $53$ & $32$ & \gd{($-$40\%)}
      & $5147$ & $2920$ & \gd{($-$43\%)} & $0.61$ & $0.29$ & \gd{($-$52\%)} & $34.3$ & $22.6$ & \gd{($-$34\%)} \\
    \bottomrule
  \end{tabular}}
\end{table}

All results in this subsection use the full $256$k context window with $w{=}8$ concurrent sessions.
Table~\ref{tab:main} compares ReFold with \strat{base} across five benchmarks and two models at this operating point.
Resolved-task counts differ by $-1$ to $+3$ out of $50$, within the run-to-run variability of the setup (Appendix~\ref{sec:appendix-metricdef}).
With task resolution remaining comparable, ReFold reduces both the total tokens processed and the memory each session requires.
It shortens trajectories in nine of the ten settings, by up to $40\%$.
Fewer steps over smaller rendered contexts reduce token usage by $18$--$53\%$ on Qwen3.6-27B and $42$--$60\%$ on Qwen3.6-35B-A3B.
Smaller maximum prompts also lower peak per-session KV-cache memory by $39$--$58\%$ and $30$--$55\%$, respectively.
The token savings reduce estimated cost by $2$--$35\%$ and $18$--$45\%$, after accounting for cached-input discounts and completion costs, while the smaller KV-cache footprint leaves room for higher concurrency.

\subsection{Performance under Resource Constraints}
\label{sec:results-scarcity}

Table~\ref{tab:scarcity} examines whether these resource savings improve execution under limited context budgets and increased serving load.
All rows use Qwen3.6-27B on SWE-bench Verified, with $50$ tasks per pair except $w{=}64$, which is read over $71$ tasks.
We vary the context budget at fixed $w{=}8$, increase concurrency at the full window, and evaluate both constraints jointly.

\begin{table}[t]
  \caption{Performance under context-budget and concurrency constraints. We vary the context budget at fixed $w{=}8$, increase concurrency at the full context window, and combine both constraints. Each pair compares the reference baseline with and without ReFold; parentheses indicate the changes. }
  \label{tab:scarcity}
  \centering \small
  \setlength{\tabcolsep}{3pt}
  \renewcommand{\arraystretch}{1.14}
  \resizebox{\linewidth}{!}{%
  \begin{tabular}{ll*{3}{P{0pt}}P{5.8pt}P{4.7pt}P{6.9pt}P{0pt}}
    \toprule
    Context, $w$ & Arm & \multicolumn{2}{c}{Resolve} & \multicolumn{2}{c}{Tokens (k)}
      & \multicolumn{2}{c}{Cost (\textcent)} & \multicolumn{2}{c}{Duration (min)} & \multicolumn{2}{c}{Compactions}
      & \multicolumn{2}{c}{KV peak (\%)} & \multicolumn{2}{c}{Queue (s)} \\
    \midrule
    $256$k, $w{=}8$ & \strat{base} & $38$ & & $965$ & & $19.5$ & & $9.8$ & & --- & & $42$ & & $0$ & \\
    \rowcolor{black!7} & \quad + ReFold & $38$ & \nd{($\pm$0)} & $505$ & \gd{($-$48\%)}
      & $12.9$ & \gd{($-$34\%)} & $9.4$ & \gd{($-$4\%)} & --- & & $29$ & \gd{($-$13)} & $0$ & \nd{($\pm$0)} \\
    \midrule
    $64$k, $w{=}8$ & \strat{compact} & $37$ & & $882$ & & $18.5$ & & $9.8$ & & $0.1$ & & $43$ & & $0$ & \\
    \rowcolor{black!7} & \quad + ReFold & $39$ & \nd{($+$2)} & $528$ & \gd{($-$40\%)}
      & $14.1$ & \gd{($-$24\%)} & $8.2$ & \gd{($-$17\%)} & $0.1$ & \nd{($\pm$0)} & $28$ & \gd{($-$15)} & $0$ & \nd{($\pm$0)} \\
    \midrule
    $32$k, $w{=}8$ & \strat{compact} & $37$ & & $664$ & & $15.3$ & & $10.7$ & & $1.0$ & & $33$ & & $0$ & \\
    \rowcolor{black!7} & \quad + ReFold & $35$ & \nd{($-$2)} & $374$ & \gd{($-$44\%)}
      & $9.8$ & \gd{($-$36\%)} & $9.3$ & \gd{($-$13\%)} & $0.1$ & \gd{($-$92\%)} & $30$ & \gd{($-$3)} & $0$ & \nd{($\pm$0)} \\
    \midrule
    $24$k, $w{=}8$ & \strat{compact} & $37$ & & $621$ & & $15.5$ & & $11.4$ & & $2.3$ & & $30$ & & $0$ & \\
    \rowcolor{black!7} & \quad + ReFold & $40$ & \nd{($+$3)} & $400$ & \gd{($-$36\%)}
      & $11.2$ & \gd{($-$28\%)} & $10.1$ & \gd{($-$12\%)} & $0.4$ & \gd{($-$83\%)} & $22$ & \gd{($-$8)} & $0$ & \nd{($\pm$0)} \\
    \midrule
    $16$k, $w{=}8$ & \strat{compact} & $38$ & & $450$ & & $13.5$ & & $11.6$ & & $5.0$ & & $22$ & & $0$ & \\
    \rowcolor{black!7} & \quad + ReFold & $37$ & \nd{($-$1)} & $391$ & \gd{($-$13\%)}
      & $13.5$ & \nd{($\pm$0\%)} & $10.0$ & \gd{($-$14\%)} & $3.3$ & \gd{($-$33\%)} & $19$ & \gd{($-$3)} & $0$ & \nd{($\pm$0)} \\
    \midrule
    $256$k, $w{=}16$ & \strat{base} & $31$ & & $1290$ & & $28.1$ & & $26.2$ & & --- & & $84$ & & $0.02$ & \\
    \rowcolor{black!7} & \quad + ReFold & $38$ & \gd{($+$7)} & $600$ & \gd{($-$53\%)}
      & $15.1$ & \gd{($-$46\%)} & $15.4$ & \gd{($-$41\%)} & --- & & $48$ & \gd{($-$36)} & $0$ & \gd{($-$0.02)} \\
    \midrule
    $256$k, $w{=}32$ & \strat{base} & $20$ & & $1853$ & & $41.7$ & & $69.5$ & & --- & & $100$ & & $1.91$ & \\
    \rowcolor{black!7} & \quad + ReFold & $35$ & \gd{($+$15)} & $623$ & \gd{($-$66\%)}
      & $19.0$ & \gd{($-$54\%)} & $46.4$ & \gd{($-$33\%)} & --- & & $84$ & \gd{($-$16)} & $0$ & \gd{($-$1.91)} \\
    \midrule
    $256$k, $w{=}48$ & \strat{base} & $7$ & & $2103$ & & $61.0$ & & $90.3$ & & --- & & $100$ & & $8.96$ & \\
    \rowcolor{black!7} & \quad + ReFold & $32$ & \gd{($+$25)} & $639$ & \gd{($-$70\%)}
      & $17.8$ & \gd{($-$71\%)} & $69.3$ & \gd{($-$23\%)} & --- & & $99$ & \nd{($-$1)} & $0.02$ & \gd{($-$8.94)} \\
    \midrule
    $256$k, $w{=}64$ & \strat{base} & $18$ & & $517$ & & $15.1$ & & $87.6$ & & --- & & $100$ & & $42.2$ & \\
    \rowcolor{black!7} & \quad + ReFold & $39$ & \gd{($+$21)} & $368$ & \gd{($-$29\%)}
      & $12.2$ & \gd{($-$19\%)} & $84.6$ & \gd{($-$3\%)} & --- & & $100$ & \nd{($\pm$0)}
      & $23.1$ & \gd{($-$19.1)} \\
    \midrule
    $16$k, $w{=}48$ & \strat{compact} & $36$ & & $459$ & & $16.6$ & & $87.3$ & & $5.4$ & & $100$ & & $0.87$ & \\
    \rowcolor{black!7} & \quad + ReFold & $36$ & \nd{($\pm$0)} & $349$ & \gd{($-$24\%)}
      & $13.7$ & \gd{($-$17\%)} & $63.3$ & \gd{($-$28\%)} & $3.4$ & \gd{($-$37\%)} & $92$ & \gd{($-$8)} & $0.02$ & \gd{($-$0.85)} \\
    \bottomrule
  \end{tabular}}
\end{table}

\paragraph{Context budget constraints.}
By removing repeated content and folding completed turns, ReFold slows context growth and delays threshold-triggered compaction.
At fixed $w{=}8$, mean cost decreases by $24$--$36\%$ and median duration by $12$--$17\%$ across the $24$k, $32$k, and $64$k budgets, with compaction reduction reaching $92\%$ at $32$k.
These savings reflect both smaller model inputs and fewer auxiliary summarization calls.
At $16$k, compactions and duration decrease by $33\%$ and $14\%$, respectively.
Resolved-task counts differ by at most three out of $50$ across all tested budgets.

\paragraph{Concurrent serving.}
The smaller per-session KV-cache footprint reported in Table~\ref{tab:main} reduces memory demand under concurrent execution, alleviating serving contention.
At $w{=}16$, $32$, and $48$, ReFold reduces token usage by $53$--$70\%$, mean cost by $46$--$71\%$, and median duration by $23$--$41\%$.
Resolved-task counts increase from $31$ to $38$, $20$ to $35$, and $7$ to $32$, respectively.
At $w{=}48$, mean scheduling delay falls from $8.96$ to $0.02$\,s, consistent with lower contention allowing more tasks to finish within the execution limits.
At $w{=}64$, both conditions saturate the KV cache, yet ReFold still reduces queue delay from $42.2$ to $23.1$\,s and increases solved tasks from $18$ to $39$ out of the $71$ tasks.

\paragraph{Joint budget and concurrency constraints.}
Under simultaneous constraints, reducing the rendered context limits both compaction overhead and concurrent KV-cache demand.
With a $16$k budget and $w{=}48$, ReFold reduces token usage by $24\%$, compactions by $37\%$, mean cost by $17\%$, and median duration by $28\%$, while both conditions solve $36$ of $50$ tasks.
Peak server KV-cache occupancy decreases from $100\%$ to $92\%$, and mean scheduling delay from $0.87$ to $0.02$\,s.
The reductions in compaction frequency and queue delay contribute to improved execution efficiency.

\subsection{Ablation Studies}
\label{sec:results-ablation}

\begin{figure}[t]
  \centering
  \includegraphics[width=\linewidth]{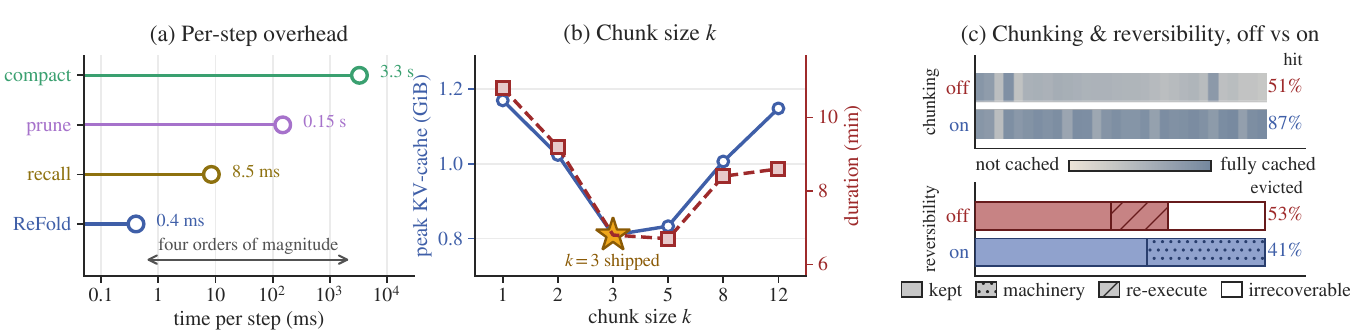}
  \setlength{\abovecaptionskip}{2pt}
  \caption{Ablation results. \textbf{(a)}~Per-step context-management overhead on a log scale. \textbf{(b)}~Peak KV-cache memory (left axis) and median duration (right axis) across chunk sizes on $20$ tasks; the axes are scaled to coincide at $k{=}3$. \textbf{(c)}~Chunking (top) and reversibility (bottom) switched off and on.}
  \label{fig:abl}
\end{figure}

All ablations use Qwen3.6-27B on SWE-bench Verified (Figure~\ref{fig:abl}): the chunk-size sweep runs $20$ tasks at the full window with $w{=}8$, and the chunking and reversibility ablations re-render recorded trajectories with each feature switched off and on.

\paragraph{Runtime overhead.}
\label{sec:results-overhead}
ReFold applies deduplication and folding during context rendering without auxiliary model calls.
Its mean rendering overhead is $0.4$\,ms per step, compared with $3.3$\,s for \strat{compact}, $0.15$\,s for \strat{prune}, and $8.5$\,ms for \strat{recall} (Figure~\ref{fig:abl}a).
ReFold therefore reduces the computational overhead of context management within the agent loop.

\paragraph{Selection of the chunk size $k$.}
\label{sec:results-recentk}
The chunk size $k$ balances context reduction and prefix-cache reuse: smaller chunks update historical context more frequently, whereas larger chunks retain more uncompressed turns.
Peak KV-cache memory and median duration are lowest at $k{=}3$--$5$ and rise toward both ends of the tested range (Figure~\ref{fig:abl}b).
With little variation in resolved-task counts, we select $k{=}3$ to limit prefix-cache invalidation while maintaining effective compression.

\paragraph{Effect of chunked rendering.}
\label{sec:results-chunking}
As in Figure~\ref{fig:sec2}e, Figure~\ref{fig:abl}c (top) traces prefix-cache reuse along one trajectory.
Rendering the context anew at every step rewrites the cached prefix almost every time it changes, which drives the hit rate down to $51\%$.
Chunked rendering confines these rewrites to boundary advances, once every $k{=}3$ steps, and keeps the context append-only in between, so the hit rate stays at $87\%$, comparable to the $84\%$ under \strat{base}.

\paragraph{Effect of reversibility.}
\label{sec:results-reversibility}
As in Figure~\ref{fig:sec2}f, Figure~\ref{fig:abl}c (bottom) classifies each evicted observation token as recoverable or irrecoverable.
Switching reversibility off disables \texttt{restore}, so removed content comes back only by re-reading an unchanged file, and even a re-executed output is compressed again once it passes the boundary instead of staying in full.
ReFold then removes $53\%$ of the observation tokens, but about a third of all observation tokens can no longer be recovered, the same share that \strat{compact} loses.
When reversibility stays on, ReFold removes $41\%$, as much as \strat{compact} evicts, yet every removed token can be restored from the history.

\section{Related Work}
\label{sec:related}

\paragraph{Context management for LLM agents.}
To sustain long-horizon agent execution, existing methods reduce interaction history through pruning or summarization.
Observation masking replaces older tool outputs with placeholders~\citep{lindenbauer2025complexitytrap}.
AgentDiet has an LLM remove useless, redundant, or expired information from earlier steps~\citep{xiao2026agentdiet}, SWE-Pruner uses a trained $0.6$B model to retain goal-relevant lines~\citep{wang2026swepruner}, and TACO evolves observation-compression rules for terminal agents~\citep{ren2026taco}.
Summarization methods also compress the agent's own messages.
Context-Folding summarizes branched sub-trajectories~\citep{sun2025contextfolding}, while AgentFold folds history at multiple scales~\citep{ye2026agentfold}.
MEM1 learns memory consolidation through reinforcement learning~\citep{zhou2025mem1}, CompactionRL jointly learns summarization and task execution~\citep{li2026compactionrl}, and SWE-MeM learns when, what, and how to compact~\citep{gao2026swemem}.
SWE-AGILE summarizes reasoning behind a sliding window~\citep{lian2026sweagile}, CAT exposes context maintenance as a tool~\citep{liu2025cat}, and ACON optimizes compression guidelines and distills the resulting compressor~\citep{kang2026acon}.
ReFold compresses both observations and completed agent turns without auxiliary summarization calls or policy training.

\paragraph{External memory and token compression.}
To mitigate the memory and computational footprints of long-context inference, existing methods broadly rely on external memory offloading or prompt-level token compression.
For external memory, early frameworks maintain historical context outside the active window.
Generative Agents retrieve memories by recency, importance, and relevance~\citep{park2023generative}; MemGPT transfers information between context and external storage~\citep{packer2023memgpt}.
A-MEM maintains linked memories~\citep{xu2025amem}, while Mem0 consolidates information in a persistent store~\citep{chhikara2025mem0}.
LCM keeps original messages in an immutable store behind a summary DAG~\citep{ehrlich2026lcm}, and ARC replaces observations with references to an addressable log~\citep{arc2026}.
For input compression, LLMLingua uses perplexity-based token filtering~\citep{jiang2023llmlingua}, LongLLMLingua incorporates question relevance~\citep{jiang2024longllmlingua}, and LLMLingua-2 uses a distilled token classifier~\citep{pan2024llmlingua2}.
At the KV cache level, SkipKV compresses key-value caches with sentence similarity~\citep{tian2026skipkv}, while Continuum retains session KV caches across tool calls~\citep{li2026continuum}.
In contrast to these lossy compaction or rigid external stores, ReFold makes every removal reversible by a restore from the preserved history, and uses chunked rendering to preserve prefix-cache locality.

\section{Conclusion}
\label{sec:conclusion}

We presented ReFold, a training-free context-rendering layer for long-horizon LLM agents.
ReFold deduplicates repeated content and folds turns marked as completed by the agent while preserving the underlying interaction history, so every removal is reversible by a restore from it.
Chunked rendering limits prefix-cache invalidation without auxiliary model calls.
Across five benchmarks and two models, ReFold reduces token usage by $18$--$60\%$, estimated cost by up to $45\%$, and peak per-session KV-cache memory by $30$--$58\%$, with no detectable reduction in task success at the full window with $w{=}8$.
Under constrained context budgets and concurrent serving, these savings reduce compaction frequency and scheduling delay, improving execution efficiency.

\subsubsection*{Reproducibility statement}
\phantomsection\addcontentsline{toc}{section}{Reproducibility statement}
ReFold is implemented as a plugin to the mini-swe-agent harness. It requires only a rendering hook
and a change to the agent's prompt that elicits the two signals, and leaves the agent's tools,
environment and interaction protocol unchanged. Section~\ref{sec:method} specifies the two operators
and their composition, and Appendix~\ref{sec:appendix-prompt} gives the exact prompt change verbatim.
Section~\ref{sec:setup} and Appendices~\ref{sec:appendix-setup} and~\ref{sec:appendix-benchmarks} list
the models, serving configuration and benchmarks, and Appendix~\ref{sec:appendix-metrics-def} defines
every reported quantity, including the rates and the cache accounting behind every cost. Every table
and figure is produced from one per-task record of every cell, and we will release the code and
these scripts.

\bibliographystyle{iclr2027_conference}
\phantomsection\addcontentsline{toc}{section}{References}
\bibliography{iclr2027_conference}

\begin{thebibliography}{39}
\providecommand{\natexlab}[1]{#1}
\providecommand{\url}[1]{\texttt{#1}}
\expandafter\ifx\csname urlstyle\endcsname\relax
  \providecommand{\doi}[1]{doi: #1}\else
  \providecommand{\doi}{doi: \begingroup \urlstyle{rm}\Url}\fi

\bibitem[Chhikara et~al.(2025)Chhikara, Khant, Aryan, Singh, and
  Yadav]{chhikara2025mem0}
Prateek Chhikara, Dev Khant, Saket Aryan, Taranjeet Singh, and Deshraj Yadav.
\newblock {Mem0}: Building production-ready {AI} agents with scalable long-term
  memory.
\newblock In \emph{European Conference on Artificial Intelligence (ECAI)},
  volume 413 of \emph{Frontiers in Artificial Intelligence and Applications},
  pp.\  2993--3000. IOS Press, 2025.

\bibitem[Chowdhury et~al.(2024)Chowdhury, Aung, Shern, Jaffe, Sherburn,
  Starace, Mays, Dias, Aljubeh, Glaese, Jimenez, Yang, Ho, Patwardhan, Liu, and
  Madry]{openai2024swebenchverified}
Neil Chowdhury, James Aung, Chan~Jun Shern, Oliver Jaffe, Dane Sherburn, Giulio
  Starace, Evan Mays, Rachel Dias, Marwan Aljubeh, Mia Glaese, Carlos~E.
  Jimenez, John Yang, Leyton Ho, Tejal Patwardhan, Kevin Liu, and Aleksander
  Madry.
\newblock Introducing {SWE-bench Verified}.
\newblock OpenAI blog,
  \url{https://openai.com/index/introducing-swe-bench-verified/}, 2024.

\bibitem[Dang et~al.(2026)Dang, Ichikawa, Fatima, and Shirahata]{arc2026}
Thang Dang, Yuma Ichikawa, Sakina Fatima, and Koichi Shirahata.
\newblock Addressable recall compaction for long context-window control in {AI}
  agents.
\newblock \emph{arXiv preprint arXiv:2607.25066}, 2026.

\bibitem[Deng et~al.(2026)Deng, Da, Pan, He, Ide, Garg, Lauffer, Park, Rane,
  Sampath, Krishnan, Kundurthy, Hendryx, Wang, Zhang, Jacobson, Liu, and
  Kenstler]{deng2025swebenchpro}
Xiang Deng, Jeff Da, Edwin Pan, Yannis~Yiming He, Charles Ide, Kanak Garg,
  Niklas Lauffer, Andrew Park, Chetan Rane, Karmini Sampath, Maya Krishnan,
  Srivatsa Kundurthy, Sean Hendryx, Zifan Wang, Chen Bo~Calvin Zhang, Noah
  Jacobson, Bing Liu, and Brad Kenstler.
\newblock {SWE-Bench Pro}: Can {AI} agents solve long-horizon software
  engineering tasks?
\newblock In \emph{International Conference on Machine Learning (ICML)}, 2026.

\bibitem[Ehrlich \& Blackman(2026)Ehrlich and Blackman]{ehrlich2026lcm}
Clint Ehrlich and Theodore Blackman.
\newblock {LCM}: Lossless context management.
\newblock \emph{arXiv preprint arXiv:2605.04050}, 2026.

\bibitem[Gao et~al.(2026)Gao, Zeng, Yu, Wangni, Wang, Cai, He, and
  Lyu]{gao2026swemem}
Shuzheng Gao, Wenhao Zeng, Zhaojian Yu, Jianqiao Wangni, Chaozheng Wang, Kai
  Cai, Shilin He, and Michael~R. Lyu.
\newblock {SWE-MeM}: Learning adaptive memory management for long-horizon
  coding agents.
\newblock \emph{arXiv preprint arXiv:2606.28434}, 2026.

\bibitem[Hsieh et~al.(2024)Hsieh, Sun, Kriman, Acharya, Rekesh, Jia, Zhang, and
  Ginsburg]{hsieh2024ruler}
Cheng-Ping Hsieh, Simeng Sun, Samuel Kriman, Shantanu Acharya, Dima Rekesh, Fei
  Jia, Yang Zhang, and Boris Ginsburg.
\newblock {RULER}: What's the real context size of your long-context language
  models?
\newblock In \emph{Conference on Language Modeling (COLM)}, 2024.

\bibitem[Jiang et~al.(2023)Jiang, Wu, Lin, Yang, and Qiu]{jiang2023llmlingua}
Huiqiang Jiang, Qianhui Wu, Chin-Yew Lin, Yuqing Yang, and Lili Qiu.
\newblock {LLMLingua}: Compressing prompts for accelerated inference of large
  language models.
\newblock In \emph{Conference on Empirical Methods in Natural Language
  Processing (EMNLP)}, 2023.

\bibitem[Jiang et~al.(2024)Jiang, Wu, Luo, Li, Lin, Yang, and
  Qiu]{jiang2024longllmlingua}
Huiqiang Jiang, Qianhui Wu, Xufang Luo, Dongsheng Li, Chin-Yew Lin, Yuqing
  Yang, and Lili Qiu.
\newblock {LongLLMLingua}: Accelerating and enhancing {LLMs} in long context
  scenarios via prompt compression.
\newblock In \emph{Annual Meeting of the Association for Computational
  Linguistics (ACL)}, 2024.

\bibitem[Jimenez et~al.(2024)Jimenez, Yang, Wettig, Yao, Pei, Press, and
  Narasimhan]{jimenez2024swebench}
Carlos~E. Jimenez, John Yang, Alexander Wettig, Shunyu Yao, Kexin Pei, Ofir
  Press, and Karthik Narasimhan.
\newblock {SWE}-bench: Can language models resolve real-world {GitHub} issues?
\newblock In \emph{International Conference on Learning Representations
  (ICLR)}, 2024.

\bibitem[Kang et~al.(2026)Kang, Chen, Han, Inan, Wutschitz, Chen, Sim, and
  Rajmohan]{kang2026acon}
Minki Kang, Wei-Ning Chen, Dongge Han, Huseyin~A. Inan, Lukas Wutschitz, Yanzhi
  Chen, Robert Sim, and Saravan Rajmohan.
\newblock {ACON}: Optimizing context compression for long-horizon {LLM} agents.
\newblock In \emph{International Conference on Machine Learning (ICML)}, 2026.

\bibitem[Li et~al.(2026{\natexlab{a}})Li, Mang, He, Zhang, Mao, Chen, Zhou,
  Cheung, Gonzalez, and Stoica]{li2026continuum}
Hanchen Li, Qiuyang Mang, Runyuan He, Qizheng Zhang, Huanzhi Mao, Xiaokun Chen,
  Hangrui Zhou, Alvin Cheung, Joseph~E. Gonzalez, and Ion Stoica.
\newblock Continuum: Efficient and robust multi-turn {LLM} agent scheduling
  with {KV} cache time-to-live.
\newblock In \emph{ICLR Workshop on Lifelong Agents: Learning, Aligning,
  Evolving}, 2026{\natexlab{a}}.

\bibitem[Li et~al.(2026{\natexlab{b}})Li, Hou, Jing, Tang, and
  Dong]{li2026compactionrl}
Yujiang Li, Zhenyu Hou, Yi~Jing, Jie Tang, and Yuxiao Dong.
\newblock {CompactionRL}: Reinforcement learning with context compaction for
  long-horizon agents.
\newblock \emph{arXiv preprint arXiv:2607.05378}, 2026{\natexlab{b}}.

\bibitem[Lian et~al.(2026)Lian, Liu, Chen, Chen, and Li]{lian2026sweagile}
Shuquan Lian, Juncheng Liu, Yazhe Chen, Yuhong Chen, and Hui Li.
\newblock {SWE-AGILE}: A software agent framework for efficiently managing
  dynamic reasoning context.
\newblock In \emph{Findings of the Association for Computational Linguistics
  (ACL Findings)}, pp.\  17536--17550, 2026.

\bibitem[Lindenbauer et~al.(2025)Lindenbauer, Slinko, Felder, Bogomolov, and
  Zharov]{lindenbauer2025complexitytrap}
Tobias Lindenbauer, Igor Slinko, Ludwig Felder, Egor Bogomolov, and Yaroslav
  Zharov.
\newblock The complexity trap: Simple observation masking is as efficient as
  {LLM} summarization for agent context management.
\newblock In \emph{NeurIPS Workshop on Deep Learning for Code (DL4C)}, 2025.

\bibitem[Liu et~al.(2026{\natexlab{a}})Liu, Qiu, Goiri, Fonseca, Bianchini, and
  Choukse]{liu2026agentic}
Banruo Liu, Haoran Qiu, {\'I}{\~n}igo Goiri, Rodrigo Fonseca, Ricardo
  Bianchini, and Esha Choukse.
\newblock Agentic coding in the wild: Characterizing {GitHub} {Copilot} traces
  at production scale.
\newblock \emph{arXiv preprint arXiv:2608.00101}, 2026{\natexlab{a}}.

\bibitem[Liu et~al.(2024)Liu, Lin, Hewitt, Paranjape, Bevilacqua, Petroni, and
  Liang]{liu2024lost}
Nelson~F. Liu, Kevin Lin, John Hewitt, Ashwin Paranjape, Michele Bevilacqua,
  Fabio Petroni, and Percy Liang.
\newblock Lost in the middle: How language models use long contexts.
\newblock \emph{Transactions of the Association for Computational Linguistics
  (TACL)}, 12:\penalty0 157--173, 2024.

\bibitem[Liu et~al.(2026{\natexlab{b}})Liu, Jiang, Yang, Li, Guo, Liu, and
  Dai]{liu2025cat}
Shukai Liu, Bo~Jiang, Jian Yang, Yizhi Li, Jinyang Guo, Xianglong Liu, and
  Bryan Dai.
\newblock Context as a tool: Context management for long-horizon {SWE}-agents.
\newblock In \emph{Findings of the Association for Computational Linguistics
  (ACL Findings)}, pp.\  20604--20617, 2026{\natexlab{b}}.

\bibitem[Merrill et~al.(2025)Merrill, Shaw, Rytting, Schmidt, and
  Konwinski]{tbench2025core}
Mike Merrill, Alex Shaw, Chris Rytting, Ludwig Schmidt, and Andy Konwinski.
\newblock {Terminal-Bench}: An evaluation framework and benchmark for agents in
  the terminal.
\newblock \url{https://www.tbench.ai/news/announcement}, 2025.

\bibitem[Merrill et~al.(2026)Merrill, Shaw, Carlini, Li, Raj, Bercovich, Shi,
  Shin, Walshe, Buchanan, Shen, Ye, Lin, Poulos, Wang, Nezhurina, Lu,
  Mastromichalakis, Xu, Chen, Liu, Zhang, Chen, Kashyap, Uslu, Li, Wu, Yan,
  Bian, Sharma, Sun, Dillmann, Anand, Lanpouthakoun, Koopah, Hu, Guha, Dreiman,
  Zhu, Krauth, Zhong, Muennighoff, Amanfu, Tan, Pimpalgaonkar, Aggarwal, Lin,
  Lan, Zhao, Liang, Wang, Wang, Zhou, Heineman, Liu, Trivedi, Yang, Lin,
  Shetty, Yang, Omi, Raoof, Li, Zhuo, Lin, Dai, Wang, Chai, Zhou, Wahdany, She,
  Hu, Dong, Zhu, Cui, Saiyed, Kolbeinsson, Hu, Rytting, Marten, Wang, Jitsev,
  Dimakis, Konwinski, and Schmidt]{merrill2026terminalbench}
Mike~A. Merrill, Alexander~G. Shaw, Nicholas Carlini, Boxuan Li, Harsh Raj,
  Ivan Bercovich, Lin Shi, Jeong~Yeon Shin, Thomas Walshe, E.~Kelly Buchanan,
  Junhong Shen, Guanghao Ye, Haowei Lin, Jason Poulos, Maoyu Wang, Marianna
  Nezhurina, Di~Lu, Orfeas~Menis Mastromichalakis, Zhiwei Xu, Zizhao Chen, Yue
  Liu, Robert Zhang, Leon~Liangyu Chen, Anurag Kashyap, Jan-Lucas Uslu, Jeffrey
  Li, Jianbo Wu, Minghao Yan, Song Bian, Vedang Sharma, Ke~Sun, Steven
  Dillmann, Akshay Anand, Andrew Lanpouthakoun, Bardia Koopah, Changran Hu,
  Etash Guha, Gabriel H.~S. Dreiman, Jiacheng Zhu, Karl Krauth, Li~Zhong,
  Niklas Muennighoff, Robert Amanfu, Shangyin Tan, Shreyas Pimpalgaonkar,
  Tushar Aggarwal, Xiangning Lin, Xin Lan, Xuandong Zhao, Yiqing Liang, Yuanli
  Wang, Zilong Wang, Changzhi Zhou, David Heineman, Hange Liu, Harsh Trivedi,
  John Yang, Junhong Lin, Manish Shetty, Michael Yang, Nabil Omi, Negin Raoof,
  Shanda Li, Terry~Yue Zhuo, Wuwei Lin, Yiwei Dai, Yuxin Wang, Wenhao Chai,
  Shang Zhou, Dariush Wahdany, Ziyu She, Jiaming Hu, Zhikang Dong, Yuxuan Zhu,
  Sasha Cui, Ahson Saiyed, Arinbj{\"o}rn Kolbeinsson, Jesse Hu,
  Christopher~Michael Rytting, Ryan Marten, Yixin Wang, Jenia Jitsev, Alex
  Dimakis, Andy Konwinski, and Ludwig Schmidt.
\newblock {Terminal-Bench}: Benchmarking agents on hard, realistic tasks in
  command line interfaces.
\newblock In \emph{International Conference on Learning Representations
  (ICLR)}, pp.\  40903--40986, 2026.

\bibitem[Packer et~al.(2023)Packer, Wooders, Lin, Fang, Patil, Stoica, and
  Gonzalez]{packer2023memgpt}
Charles Packer, Sarah Wooders, Kevin Lin, Vivian Fang, Shishir~G. Patil, Ion
  Stoica, and Joseph~E. Gonzalez.
\newblock {MemGPT}: Towards {LLMs} as operating systems.
\newblock \emph{arXiv preprint arXiv:2310.08560}, 2023.

\bibitem[Pan et~al.(2024)Pan, Wu, Jiang, Xia, Luo, Zhang, Lin, R{\"u}hle, Yang,
  Lin, Zhao, Qiu, and Zhang]{pan2024llmlingua2}
Zhuoshi Pan, Qianhui Wu, Huiqiang Jiang, Menglin Xia, Xufang Luo, Jue Zhang,
  Qingwei Lin, Victor R{\"u}hle, Yuqing Yang, Chin-Yew Lin, H.~Vicky Zhao, Lili
  Qiu, and Dongmei Zhang.
\newblock {LLMLingua-2}: Data distillation for efficient and faithful
  task-agnostic prompt compression.
\newblock In \emph{Findings of the Association for Computational Linguistics
  (ACL Findings)}, 2024.

\bibitem[Park et~al.(2023)Park, O'Brien, Cai, Morris, Liang, and
  Bernstein]{park2023generative}
Joon~Sung Park, Joseph~C. O'Brien, Carrie~J. Cai, Meredith~Ringel Morris, Percy
  Liang, and Michael~S. Bernstein.
\newblock Generative agents: Interactive simulacra of human behavior.
\newblock In \emph{ACM Symposium on User Interface Software and Technology
  (UIST)}, 2023.

\bibitem[Ren et~al.(2026)Ren, Wu, Li, Zhu, Xu, Feng, Yuan, Zhang,
  Batista-Navarro, Yang, and Lin]{ren2026taco}
Jincheng Ren, Siwei Wu, Yizhi Li, Kang Zhu, Shu Xu, Boyu Feng, Ruibin Yuan, Wei
  Zhang, Riza Batista-Navarro, Jian Yang, and Chenghua Lin.
\newblock A self-evolving framework for efficient terminal agents via
  observational context compression.
\newblock \emph{arXiv preprint arXiv:2604.19572}, 2026.

\bibitem[Schick et~al.(2023)Schick, Dwivedi-Yu, Dess{\`i}, Raileanu, Lomeli,
  Hambro, Zettlemoyer, Cancedda, and Scialom]{schick2023toolformer}
Timo Schick, Jane Dwivedi-Yu, Roberto Dess{\`i}, Roberta Raileanu, Maria
  Lomeli, Eric Hambro, Luke Zettlemoyer, Nicola Cancedda, and Thomas Scialom.
\newblock Toolformer: Language models can teach themselves to use tools.
\newblock In \emph{Advances in Neural Information Processing Systems
  (NeurIPS)}, volume~36, pp.\  68539--68551, 2023.

\bibitem[Shinn et~al.(2023)Shinn, Cassano, Gopinath, Narasimhan, and
  Yao]{shinn2023reflexion}
Noah Shinn, Federico Cassano, Ashwin Gopinath, Karthik Narasimhan, and Shunyu
  Yao.
\newblock Reflexion: Language agents with verbal reinforcement learning.
\newblock In \emph{Advances in Neural Information Processing Systems
  (NeurIPS)}, volume~36, pp.\  8634--8652, 2023.

\bibitem[Sun et~al.(2026)Sun, Lu, Ling, Liu, Yao, Yang, and
  Chen]{sun2025contextfolding}
Weiwei Sun, Miao Lu, Zhan Ling, Kang Liu, Xuesong Yao, Yiming Yang, and Jiecao
  Chen.
\newblock Scaling long-horizon agent via context folding.
\newblock In \emph{International Conference on Machine Learning (ICML)}, 2026.

\bibitem[{The Terminal-Bench Team}(2026)]{tbench2026v21}
{The Terminal-Bench Team}.
\newblock {Terminal-Bench} 2.1.
\newblock \url{https://www.tbench.ai/news/terminal-bench-2-1}, 2026.

\bibitem[Tian et~al.(2026)Tian, Azizi, Zhao, Potraghloo, McPherson, Sridhar,
  Wang, Zhang, Pedram, and Kundu]{tian2026skipkv}
Jiayi Tian, Seyedarmin Azizi, Yequan Zhao, Erfan~B Potraghloo, Sean McPherson,
  Sharath~N Sridhar, Zhengyang Wang, Zheng Zhang, Massoud Pedram, and Souvik
  Kundu.
\newblock {SkipKV}: Selective skipping of {KV} generation and storage for
  efficient inference with large reasoning models.
\newblock \emph{Proceedings of Machine Learning and Systems}, 8:\penalty0
  1496--1514, 2026.

\bibitem[Wang et~al.(2025)Wang, Li, Song, Xu, Tang, Zhuge, Pan, Song, Li,
  Singh, Tran, Li, Ma, Zheng, Qian, Shao, Muennighoff, Zhang, Hui, Lin,
  Brennan, Peng, Ji, and Neubig]{wang2024openhands}
Xingyao Wang, Boxuan Li, Yufan Song, Frank~F. Xu, Xiangru Tang, Mingchen Zhuge,
  Jiayi Pan, Yueqi Song, Bowen Li, Jaskirat Singh, Hoang~H. Tran, Fuqiang Li,
  Ren Ma, Mingzhang Zheng, Bill Qian, Yanjun Shao, Niklas Muennighoff, Yizhe
  Zhang, Binyuan Hui, Junyang Lin, Robert Brennan, Hao Peng, Heng Ji, and
  Graham Neubig.
\newblock {OpenHands}: An open platform for {AI} software developers as
  generalist agents.
\newblock In \emph{International Conference on Learning Representations
  (ICLR)}, 2025.

\bibitem[Wang et~al.(2026)Wang, Shi, Yang, Zhang, He, Lian, Chen, Ye, Cai, and
  Gu]{wang2026swepruner}
Yuhang Wang, Yuling Shi, Mo~Yang, Rongrui Zhang, Shilin He, Heng Lian, Yuting
  Chen, Siyu Ye, Kai Cai, and Xiaodong Gu.
\newblock {SWE-Pruner}: Self-adaptive context pruning for coding agents.
\newblock \emph{arXiv preprint arXiv:2601.16746}, 2026.

\bibitem[Xia et~al.(2025)Xia, Deng, Dunn, and Zhang]{xia2024agentless}
Chunqiu~Steven Xia, Yinlin Deng, Soren Dunn, and Lingming Zhang.
\newblock Demystifying {LLM}-based software engineering agents.
\newblock \emph{Proceedings of the ACM on Software Engineering}, 2\penalty0
  (FSE), 2025.
\newblock Article FSE037.

\bibitem[Xiao et~al.(2026)Xiao, Gao, Peng, and Xiong]{xiao2026agentdiet}
Yuan-An Xiao, Pengfei Gao, Chao Peng, and Yingfei Xiong.
\newblock Reducing cost of {LLM} agents with trajectory reduction.
\newblock \emph{Proceedings of the ACM on Software Engineering}, 3\penalty0
  (FSE):\penalty0 1241--1263, 2026.

\bibitem[Xu et~al.(2025)Xu, Liang, Mei, Gao, Tan, and Zhang]{xu2025amem}
Wujiang Xu, Zujie Liang, Kai Mei, Hang Gao, Juntao Tan, and Yongfeng Zhang.
\newblock {A-Mem}: Agentic memory for {LLM} agents.
\newblock In \emph{Advances in Neural Information Processing Systems
  (NeurIPS)}, volume~38, pp.\  17577--17604, 2025.

\bibitem[Yang et~al.(2024)Yang, Jimenez, Wettig, Lieret, Yao, Narasimhan, and
  Press]{yang2024sweagent}
John Yang, Carlos~E. Jimenez, Alexander Wettig, Kilian Lieret, Shunyu Yao,
  Karthik Narasimhan, and Ofir Press.
\newblock {SWE}-agent: Agent-computer interfaces enable automated software
  engineering.
\newblock In \emph{Advances in Neural Information Processing Systems
  (NeurIPS)}, volume~37, pp.\  50528--50652, 2024.

\bibitem[Yao et~al.(2023)Yao, Zhao, Yu, Du, Shafran, Narasimhan, and
  Cao]{yao2023react}
Shunyu Yao, Jeffrey Zhao, Dian Yu, Nan Du, Izhak Shafran, Karthik Narasimhan,
  and Yuan Cao.
\newblock {ReAct}: Synergizing reasoning and acting in language models.
\newblock In \emph{International Conference on Learning Representations
  (ICLR)}, 2023.

\bibitem[Ye et~al.(2026)Ye, Zhang, Li, Yin, Tao, Zhao, Su, Zhang, Qiao, Wang,
  Xie, Huang, Zhou, Chen, and Jiang]{ye2026agentfold}
Rui Ye, Zhongwang Zhang, Kuan Li, Huifeng Yin, Zhengwei Tao, Yida Zhao,
  Liangcai Su, Liwen Zhang, Zile Qiao, Xinyu Wang, Pengjun Xie, Fei Huang,
  Jingren Zhou, Siheng Chen, and Yong Jiang.
\newblock {AgentFold}: Long-horizon web agents with proactive context folding.
\newblock In \emph{International Conference on Learning Representations
  (ICLR)}, 2026.

\bibitem[Zan et~al.(2025)Zan, Huang, Liu, Chen, Xin, Zhang, Liu, Li, Chen,
  Zhong, Liu, Xiao, Chen, Zhang, Su, Liu, Long, Ding, and
  Xiang]{zan2025multiswebench}
Daoguang Zan, Zhirong Huang, Wei Liu, Hanwu Chen, Shulin Xin, Linhao Zhang,
  Qi~Liu, Aoyan Li, Lu~Chen, Xiaojian Zhong, Siyao Liu, Yongsheng Xiao,
  Liangqiang Chen, Yuyu Zhang, Jing Su, Tianyu Liu, Rui Long, Ming Ding, and
  Liang Xiang.
\newblock {Multi-SWE-bench}: A multilingual benchmark for issue resolving.
\newblock In \emph{Advances in Neural Information Processing Systems (NeurIPS),
  Datasets and Benchmarks Track}, volume~38, 2025.

\bibitem[Zhou et~al.(2026)Zhou, Qu, Wu, Kim, Prakash, Rus, Low, and
  Liang]{zhou2025mem1}
Zijian Zhou, Ao~Qu, Zhaoxuan Wu, Sunghwan Kim, Alok Prakash, Daniela Rus, Bryan
  Kian~Hsiang Low, and Paul~Pu Liang.
\newblock {MEM1}: Learning to synergize memory and reasoning for efficient
  long-horizon agents.
\newblock In \emph{International Conference on Learning Representations
  (ICLR)}, 2026.

\end{thebibliography}

\appendix
\clearpage
{\hypersetup{linktoc=all}%
 \setcounter{tocdepth}{2}%
 \makeatletter\patchcmd{\l@section}{1.0em}{0.5em}{}{}\makeatother %
 \tableofcontents}
\clearpage
\setlength{\belowcaptionskip}{4pt}

\section{Experimental setup and measurement conventions}
\label{sec:appendix-metricdef}

\subsection{Setup}
\label{sec:appendix-setup}

Every run in this paper serves its model on two NVIDIA RTX 6000 Ada GPUs with tensor parallelism $2$,
and every arm runs in the same mini-swe-agent harness with the same tools and execution environment.
Table~\ref{tab:setup} lists the settings every run shares.

\begin{table}[h]
  \caption{Serving, decoding and harness settings shared by every run.}
  \label{tab:setup}
  \centering \small
  \renewcommand{\arraystretch}{1.1}
  \begin{tabular}{@{}ll@{}}
    \toprule
    Setting & Value \\
    \midrule
    \rowcolor{black!5}\multicolumn{2}{@{}l}{\textsc{Serving}} \\
    Models & Qwen3.6-27B (dense), Qwen3.6-35B-A3B (MoE) \\
    GPUs & $2\times$ NVIDIA RTX 6000 Ada ($48$\,GB), tensor parallelism $2$ \\
    Inference engine & vLLM 0.19.1, bf16, prefix caching enabled \\
    GPU memory utilization & $0.92$ \\
    Max.\ running sequences & $128$ \\
    Max.\ batched tokens per iteration & $16{,}384$ \\
    KV-cache pool & $31.3$\,GiB, $513$k tokens (27B); $16.9$\,GiB, $887$k tokens (A3B) \\
    Server restart & before every run, starting from an empty prefix cache \\
    \midrule
    \rowcolor{black!5}\multicolumn{2}{@{}l}{\textsc{Decoding}} \\
    Temperature & $0$ \\
    Max.\ generated tokens per step & $16{,}384$ \\
    Context window & $262{,}144$ tokens \\
    Usable context & context window minus $16{,}384$, which is $245{,}760$ at the full window \\
    Capped budgets & $8$k, $16$k, $24$k, $32$k and $64$k tokens of usable context \\
    \midrule
    \rowcolor{black!5}\multicolumn{2}{@{}l}{\textsc{Harness}} \\
    Agent & mini-swe-agent, ReAct-style, one bash tool \\
    Step limit & $250$ model calls per task \\
    Request timeout & a request with no token for $600$\,s ends the task, without retry \\
    Observation length & truncated to $10{,}000$ characters \\
    Concurrent sessions $w$ & $8$ per server unless stated otherwise \\
    Tasks & $50$ per benchmark, stratified by repository and difficulty \\
    \bottomrule
  \end{tabular}
\end{table}

\subsection{Benchmarks}
\label{sec:appendix-benchmarks}

Table~\ref{tab:benchmarks} describes the five benchmarks. Each contributes $50$ tasks, sampled with
stratification by repository and difficulty, and all arms of a comparison run on the same tasks. In the
three SWE-bench variants the agent receives a repository and an issue, and its final patch counts as a fix
when the tests that reproduce the issue pass and the tests that passed before still pass. In the two
Terminal-Bench releases the agent receives a task in a container, and the task's tests check the state it
leaves behind. Every task runs in its benchmark's official container and is scored by the official evaluator.

\begin{table}[ht]
  \caption{The five benchmarks. Our $50$ tasks cover every repository of the SWE-bench Pro public set and
    seven languages of Multi-SWE-bench, so the results are not tied to Python repositories.}
  \label{tab:benchmarks}
  \centering \footnotesize
  \setlength{\tabcolsep}{4pt}
  \renewcommand{\arraystretch}{1.15}
  \begin{tabular}{@{}l >{\raggedright\arraybackslash}p{5.2cm} >{\raggedright\arraybackslash}p{1.75cm}
                  >{\raggedright\arraybackslash}p{3.45cm}@{}}
    \toprule
    Benchmark & Task & Full set & Our $50$ tasks \\
    \midrule
    SWE-bench Verified & GitHub issues in Python repositories, the human-validated subset of SWE-bench
      & $500$ tasks & $8$ repositories, all Python \\
    SWE-bench Pro & Long-horizon issues in application and developer-tool repositories, reference patches
      of $107$ lines over $4.1$ files on average & $731$ tasks (public set) & all $11$ repositories; Go $19$,
      Python $14$, TypeScript $12$, JavaScript $5$ \\
    Multi-SWE-bench & GitHub issues in Java, TypeScript, JavaScript, Go, Rust, C and C++ repositories
      & $1{,}632$ tasks & $19$ repositories; Rust $13$, C++ $10$, JavaScript $9$, C $6$, Java $5$,
      TypeScript $4$, Go $3$ \\
    Terminal-Bench 1.0 & Tasks in a Linux container through the terminal: scientific workflows, networking,
      games, data analysis, security & $80$ tasks & $50$ of the $80$ \\
    Terminal-Bench 2.1 & Terminal-Bench 2.0, a harder and better verified set, with $28$ of its tasks fixed
      & $89$ tasks & $50$ of the $89$ \\
    \bottomrule
  \end{tabular}
\end{table}

\subsection{Metrics}
\label{sec:appendix-metrics-def}

Table~\ref{tab:metrics} defines every metric. For task $i$ of the $N$ in a cell, model call $j$ sends
$p_{ij}$ prompt tokens, of which $c_{ij}$ are served from the prefix cache, and receives $g_{ij}$
completion tokens, and a call an arm makes for itself, such as a summary, counts like any other.
Token and cost figures are means over tasks, and step, duration and KV-cache figures are medians. A median
of step counts is rounded to the nearest integer, with halves rounded up.
Costs use hosted rates representative of August 2026, per million tokens $\pi_u{=}\$0.30$ uncached,
$\pi_c{=}\$0.15$ cached and $\pi_g{=}\$2.70$ completion for Qwen3.6-27B, and $\$0.10$, $\$0.05$ and
$\$0.90$ for Qwen3.6-35B-A3B. In the KV-cache formula, $\kappa$ is the cache
per token of the full-attention layers at bf16, $64$\,KiB for the 27B and $20$\,KiB for the A3B, and
$s$ the fixed state of the linear-attention layers, $0.14$ and $0.06$\,GiB. Server-side metrics are
read from vLLM's counters over the $R$ requests of a run, where request $r$ arrives at
$t^{\mathrm{arr}}_r$ and is scheduled at $t^{\mathrm{sch}}_r$.

\begin{table}[ht]
  \caption{Definitions of the metrics used in every table and figure.}
  \label{tab:metrics}
  \centering \footnotesize
  \setlength{\tabcolsep}{4pt}
  \renewcommand{\arraystretch}{1.3}
  \begin{tabular}{@{}l >{\raggedright\arraybackslash}p{5.6cm} l@{}}
    \toprule
    Metric & Meaning & Formula \\
    \midrule
    \rowcolor{black!5}\multicolumn{3}{@{}l}{\textsc{Per task, from the agent}} \\
    Resolve & tasks the official evaluator scores as solved, out of $N$ & $\sum_i \mathbf{1}[\text{task } i \text{ solved}]$ \\
    Exhausted & tasks with a request longer than the usable context $B$ & $\sum_i \mathbf{1}[\exists j{:}\ p_{ij} > B]$ \\
    Timeouts & tasks ended by a model request that streams no token for $600$\,s & $\sum_i \mathbf{1}[\text{task } i \text{ timed out}]$ \\
    Steps & model calls of a task, median over tasks rounded to an integer & $\operatorname{med}_i n_i$ \\
    Tokens & prompt and completion tokens of a task & $\frac{1}{N}\sum_i\sum_j (p_{ij}+g_{ij})$ \\
    Cost & tokens at hosted rates, in US cents & $\frac{1}{N}\sum_i\sum_j \bigl(\pi_u(p_{ij}{-}c_{ij}) + \pi_c c_{ij} + \pi_g g_{ij}\bigr)$ \\
    Cache hit & share of prompt tokens served from the prefix cache & $\sum_{i,j} c_{ij} \,/\, \sum_{i,j} p_{ij}$ \\
    Duration & wall clock from a task's first request to its last response & $\operatorname{med}_i \bigl(t^{\mathrm{end}}_i - t^{\mathrm{start}}_i\bigr)$ \\
    KV cache & memory one session holds at its largest prompt, over both GPUs & $\operatorname{med}_i \bigl(\kappa \max_j p_{ij} + s\bigr)$ \\
    Compactions & summarize-and-restart events of \strat{compact} & $\frac{1}{N}\sum_i m_i$ \\
    \midrule
    \rowcolor{black!5}\multicolumn{3}{@{}l}{\textsc{Per run, from the server}} \\
    KV peak & largest share of the KV-cache pool in use & $\max_t u(t) / U$ \\
    Queue & mean wait before the scheduler admits a request & $\frac{1}{R}\sum_r \bigl(t^{\mathrm{sch}}_r - t^{\mathrm{arr}}_r\bigr)$ \\
    Preemptions & requests evicted from the KV cache and later recomputed & count \\
    \bottomrule
  \end{tabular}
\end{table}

\paragraph{Reading the tables.}
In paired rows ReFold sits under the arm it is compared with, and a bracketed figure is its change from
that arm, green when better, red when worse and grey when it is zero or within run-to-run noise.

\subsection{Run-to-run variability}
\label{sec:appendix-variability}

Two runs of one configuration do not produce the same trajectories, since batched decoding at
temperature $0$ is not deterministic. We ran seven configurations twice on the same tasks
(Table~\ref{tab:aa}). In six of the seven pairs the resolve count moves by $1$ to $3$ of $50$, and $2$ to $7$ tasks
flip in each direction, so a difference of up to four tasks between two arms is within run-to-run noise. \strat{mask}, which runs
closest to the context limit, moves by $8$.

\begin{table}[ht]
  \caption{Seven configurations run twice on the same tasks. Lost and gained are the tasks solved
    only in the first or only in the second run, and six of the seven pairs differ by $1$ to $3$ tasks.}
  \label{tab:aa}
  \centering \small
  \begin{tabular}{llrrrr}
    \toprule
    Arm & Context & Run 1 & Run 2 & Lost & Gained \\
    \midrule
    \strat{compact} & $16$k & $38$ & $35$ & $5$ & $2$ \\
    \strat{recall} & $16$k & $27$ & $24$ & $7$ & $4$ \\
    \strat{mask} & $16$k & $17$ & $25$ & $5$ & $13$ \\
    \strat{compact} & $64$k & $38$ & $37$ & $3$ & $2$ \\
    \strat{compact} + ReFold & $64$k & $37$ & $39$ & $2$ & $4$ \\
    \strat{base} & $256$k & $37$ & $39$ & $3$ & $5$ \\
    ReFold & $256$k & $39$ & $37$ & $5$ & $3$ \\
    \bottomrule
  \end{tabular}
\end{table}

\FloatBarrier

\section{Motivational study details}
\label{sec:appendix-motivation}

\subsection{The six strategies}
\label{sec:appendix-strategies}

Table~\ref{tab:strategies} summarizes the six strategies of Section~\ref{sec:motivation}, with the
part of the history each acts on, what decides a removal and what is left in its place. All six run in
the same harness on the same $50$ SWE-bench Verified tasks at a $16$k usable-context budget with
$w{=}8$, each configurable one at its best-resolving setting in Table~\ref{tab:dose}, and every model
call a strategy makes for itself counts toward its tokens, cost and duration. \strat{compact} here is the
second of its two $16$k runs in Table~\ref{tab:aa}, the one that logs the timing of its summary calls, and
Table~\ref{tab:scarcity} reports the first.

\begin{table}[ht]
  \caption{The six strategies of Section~\ref{sec:motivation}, by what they act on and what decides a removal. Only \strat{compact} touches the agent's own messages and keeps every task within the budget.}
  \label{tab:strategies}
  \centering \footnotesize
  \setlength{\tabcolsep}{4pt}
  \renewcommand{\arraystretch}{1.15}
  \begin{tabular}{@{}l >{\raggedright\arraybackslash}p{2.5cm} >{\raggedright\arraybackslash}p{3.1cm} >{\raggedright\arraybackslash}p{2.8cm} >{\raggedright\arraybackslash}p{3.1cm}@{}}
    \toprule
    Strategy & Acts on & Removal decided by & Left in its place & As in \\
    \midrule
    \strat{base}    & nothing & --- & --- & harness default \\
    \strat{clip}    & observations older than the last three & a $1{,}000$-character cap & a truncation marker
      & harness default \\
    \strat{mask}    & whole observation & recency: older than the last $N{=}10$
      & a one-line placeholder & observation masking \citep{lindenbauer2025complexitytrap} \\
    \strat{prune}   & lines within observations & a trained $0.6$B scorer & the lines it keeps
      & SWE-Pruner \citep{wang2026swepruner} \\
    \strat{recall}  & whole observations & age, once the context passes $75\%$ of the budget
      & an addressable placeholder; \texttt{recall <id>} brings it back & ARC \citep{arc2026} \\
    \strat{compact} & the whole history, including agent messages
      & an extra model call, once the context passes $75\%$ of the budget
      & a generated summary & Claude Code, Codex \\
    \bottomrule
  \end{tabular}
\end{table}

\paragraph{\strat{base}.}
The harness default, which sends the full interaction history at every step.

\paragraph{\strat{clip}.}
Every observation older than the three most recent is cut to a fixed number of characters, keeping
its first two thirds and its last third around a truncation marker. The cap is $1{,}000$ characters.

\paragraph{\strat{mask}.}
Observation masking \citep{lindenbauer2025complexitytrap}. Tool observations older than the $N{=}10$
most recent are replaced by a one-line placeholder, and the agent's reasoning and commands stay in
place.

\paragraph{\strat{prune}.}
SWE-Pruner \citep{wang2026swepruner} with the $0.6$B scorer released with it. The agent attaches a
focus question to each command, and the scorer keeps
the lines of the output that are relevant to that question at a threshold of $0.5$. Only file reads
and searches are pruned, outputs shorter than $500$ characters are left intact, and an output still
longer than $10{,}000$ characters after pruning keeps its first and last $5{,}000$.

\paragraph{\strat{recall}.}
Addressable eviction modeled on ARC \citep{arc2026}. Once the estimated prompt passes $75\%$ of
the budget, the oldest tool observations of at least $500$ characters move to an append-only store
until the prompt falls below $60\%$, while the six most recent observations stay, fewer when that is
not enough. Each evicted observation leaves a placeholder with its index, command and size, and the
command \texttt{recall <id>} returns the stored text verbatim without running anything. Its trigger matches that of \strat{compact}.

\paragraph{\strat{compact}.}
Threshold-triggered summarization in the style of Claude Code and Codex. Once the
estimated prompt passes $75\%$ of the budget, the oldest $60\%$ of the history not yet summarized is
sent to the same served model, which returns a summary of at most $\max(2048, B/8)$ tokens for a
budget of $B$ tokens, and the summary replaces that history for good. The six most recent
observations are kept, fewer when they do not fit.

\subsection{Hyperparameter settings}
\label{sec:appendix-tuning}

\strat{clip} and \strat{mask} each expose one knob that sets how much they remove, the character
cap and the window size. Table~\ref{tab:dose} reports each at three settings next to \strat{base} and
\strat{compact}, with the mean tokens per task and the median duration in minutes, and Section~\ref{sec:motivation} uses the setting that resolves the most, marked in bold, a
cap of $1{,}000$ characters and $N{=}10$. \strat{prune}, \strat{recall} and \strat{compact} run at the
settings described above.

\begin{table}[ht]
  \caption{\strat{clip} and \strat{mask} at three settings of their knob at a $16$k budget. Each resolves the most at its tightest setting, still under half of what \strat{compact} resolves.}
  \label{tab:dose}
  \centering \small
  \setlength{\tabcolsep}{4.5pt}
  \renewcommand{\arraystretch}{1.12}
  \begin{tabular}{llrrrrrr}
    \toprule
    Arm & Setting & Resolve & Exhausted & Tokens (k) & KV cache (GiB) & Cache hit & Dur. \\
    \midrule
    \strat{base} & --- & $4$ & $43$ & $288$ & $1.12$ & $87.4\%$ & $5.2$ \\
    \midrule
    \strat{clip} & cap $4000$ & $7$ & $41$ & $302$ & $1.11$ & $86.0\%$ & $6.0$ \\
    & cap $2000$ & $6$ & $42$ & $325$ & $1.12$ & $82.4\%$ & $6.9$ \\
    & cap $1000$ & $\mathbf{13}$ & $33$ & $358$ & $1.11$ & $77.1\%$ & $8.0$ \\
    \midrule
    \strat{mask} & $N{=}40$ & $4$ & $45$ & $310$ & $1.12$ & $80.8\%$ & $6.1$ \\
    & $N{=}20$ & $13$ & $31$ & $410$ & $1.09$ & $30.0\%$ & $15.5$ \\
    & $N{=}10$ & $\mathbf{17}$ & $26$ & $483$ & $1.07$ & $15.2\%$ & $24.2$ \\
    \midrule
    \strat{compact} & --- & $35$ & $0$ & $398$ & $0.83$ & $80.4\%$ & $10.8$ \\
    \bottomrule
  \end{tabular}
\end{table}

\subsection{How Figure~\ref{fig:sec2} is measured}
\label{sec:appendix-fig2}

\paragraph{Footprint and composition (a, b).}
Panel (a) plots, at each step, the median over the tasks still running. Panel (b) splits each task's peak KV cache by the
system prompt and task, the agent's messages, and the tool observations, and scales the mean shares by the median peak.
Both panels plot only the part of the KV cache that grows with the prompt.

\paragraph{Cost and time (c, d).}
Tokens and cents are means over the $50$ tasks and durations are medians. The part of each bar
that a strategy spends on its own work is shown separately. For \strat{compact} it is the summary calls, priced like any other call. The lost cache is the prompt
tokens a strategy sends uncached beyond the hit rate of \strat{base}, priced at the difference between the
uncached and the cached rate. For time it is the wall clock between a tool result and the next
request of the agent, summed over the steps where the strategy's own machinery runs, the scorer for
\strat{prune}, the store lookup for \strat{recall}, and the summary call for \strat{compact}.

\paragraph{Prefix-cache trace (e).}
The cached share of each prompt over the first $30$ model calls of one task,
\texttt{scikit-learn\_\_scikit-learn-25931}, taken from that task's trajectory in each strategy's
run.

\paragraph{Fate of removed observation tokens (f).}
Over every prompt a strategy sends, each observation token it has removed is classed as recoverable
by the strategy's own mechanism, as reproducible by re-executing a read of a file that has not changed since, or
as irrecoverable, either because the output did not come from a file read or because the file has
changed. A token counts once for every prompt it is missing from.

\Needspace{20\baselineskip}
\section{Method details}
\label{sec:appendix-method}

\subsection{Algorithm}
\label{sec:appendix-algorithm}

Algorithm~\ref{alg:render} gives the two procedures that ReFold adds to the agent loop.

\begin{algorithm}[H]
\caption{ReFold. \textsc{Observe} runs once per step, and
         \textsc{Render} runs before each model call.}
\label{alg:render}
\small
\begin{enumerate}[leftmargin=2.2em,itemsep=0.6pt,topsep=2pt,parsep=0pt,
                  label=\footnotesize\arabic*:]
  \item \textbf{state} history $\mat{h}$ (Eq.~\ref{eq:history});\;
        markings $\textsc{progress}_j$, $\textsc{stale}_j$
        for each completed step $j$ (Eq.~\ref{eq:finished})
  \item \textbf{procedure} \textsc{Observe}$(t,\,\mat{a}_t,\,\mat{o}_t)$
  \item \quad append $\mat{a}_t \oplus \mat{o}_t$ to $\mat{h}$;\;
        extract $\textsc{progress}_t$ and $\textsc{stale}_t$ from
        $\mat{a}_t$ (Eq.~\ref{eq:prepend})
  \item \textbf{procedure} \textsc{Render}$(t)$ (Eq.~\ref{eq:render})
  \item \quad compute $b_t$ (Eq.~\ref{eq:boundary}) and
        $S_t$
  \item \quad $\mat{c} \gets
        \mat{p}_{\text{sys}} \oplus \mat{p}_{\text{task}}$
  \item \quad \textbf{for} $i = 1,\dots,t$ \textbf{do}
  \item \quad\quad \textbf{if} $i \le b_t$ and $i \in S_t$:
        append $f_i$ (Eq.~\ref{eq:fold-note})
  \item \quad\quad \textbf{else if} $i > b_t$:
        $\mat{c} \gets \mat{c} \oplus \mat{a}_i \oplus \mat{o}_i$
  \item \quad\quad \textbf{else}:
        $\mat{c} \gets \mat{c} \oplus \mat{a}_i \oplus d_i$
        (Eq.~\ref{eq:dedup})
  \item \quad \textbf{return} $\mat{c}$
\end{enumerate}
\end{algorithm}

\subsection{The prompt layer}
\label{sec:appendix-prompt}

ReFold changes the agent's prompt in four places and leaves the tools unchanged. The text below is
for $k{=}3$. First, one paragraph is appended to the system prompt. It tells
the agent that a \texttt{[context-manager]} note is harness bookkeeping rather than a shell error, and
how to see omitted content again.

\begin{promptbox}{Appended to the system prompt}
Context management: to save tokens, the harness omits parts of earlier tool outputs that already appear elsewhere and folds steps you have marked STALE. Every omission is marked by a `[context-manager]` note, which is NOT a shell error: it says what was omitted and where the same content still is. The bash command `restore <step>` shows a step's output again exactly as first shown, without running anything; re-running the original command shows its current state instead. Trust the provided context and proceed; restore or re-run only when the omitted output is necessary for your next step, since each one adds a step and token tax.
\end{promptbox}

Second, the first item of the official response-format list, which by default asks for an open-ended
reasoning section, is replaced by the two-line protocol that supplies the step progress and the
finished-step indices the operators consume. The rest of the template is kept verbatim.

\begin{promptbox}{Replaces the first response-format item}
Begin every response with these two lines, then your reasoning and your bash tool call:
  PROGRESS: <one short line on what THIS step does>
  STALE: <step numbers whose output you will not read again, or `-` if none>
Each earlier tool output is labeled `[step N]`.
STALE is about OUTPUT you are finished reading, not about the action. Name a step once you have taken what you needed from its output and will not consult that text again: you read a file and drew your conclusion, a search led nowhere, a check ruled a guess out.
Do NOT name a step whose output you will still consult: the file you are about to change, the test results you are working against, the read that pinned down the root cause. Never name the step you are writing right now, it has no result yet.
Name steps AS SOON AS you are done with them. Only the last 3 steps before this one can still be named; older numbers are ignored, so saving up a big cleanup for later is wasted.
Named steps are folded into a `[context-manager]` note that keeps their PROGRESS lines and a signature skeleton. To see a folded or omitted output again, run `restore <step>` as your bash command (for example `restore 7` or `restore 3-5`): it shows the output exactly as first shown and runs nothing. Re-run the original command instead when you need its current state.
Example first lines:
  PROGRESS: reading serializer.py to find the encode path
  STALE: 3, 4
These two lines are metadata, never let them replace your bash tool call.
\end{promptbox}

Third, every rendered tool output is prefixed with its step index, \texttt{[step $N$]}, so that steps
can be named. The harness counts from $0$, so turn $\mat{z}_i$ of Section~\ref{sec:method} carries
\texttt{[step $i{-}1$]}, and the step numbers below are these labels. Fourth, when a reply omits the \textsc{progress} line, a reminder is appended to the
newest tool output, in a firmer form after repeated misses.

\begin{promptbox}{Appended to the newest tool output after a reply without the two lines}
[context-manager] Your last response did not start with the two metadata lines. Start this one with them, then continue:
  PROGRESS: <one short line on what THIS step does>
  STALE: <step numbers you are done with, or `-`>

[context-manager] FORMAT VIOLATION (<n> in a row): every response MUST start with the PROGRESS and STALE lines. Write them FIRST, before anything else, right now.
\end{promptbox}

\subsection{Worked example}
\label{sec:appendix-example}

The boxes below follow one ReFold trajectory of Qwen3.6-27B on SWE-bench Verified with $k{=}3$,
\texttt{pydata\_\_xarray-2905}, with notes in the format of Section~\ref{sec:rendering}. Long lines are
wrapped and long outputs shortened. The green box gives the task and the amber box a reply of the agent,
blue boxes hold what ReFold renders in place of an output, and the grey box an output shown in full.

\paragraph{The task.}
Assigning an object that has a \texttt{values} attribute into an array stores that attribute instead of
the object.

\begin{taskbox}{Task of \texttt{pydata\_\_xarray-2905}, the issue shortened}
Variable.__setitem__ coercing types on objects with a values property
[...]
import xarray as xr

good_indexed, bad_indexed = xr.DataArray([None]), xr.DataArray([None])

class HasValues(object):
    values = 5

good_indexed.loc[{'dim_0': 0}] = set()
bad_indexed.loc[{'dim_0': 0}] = HasValues()

# correct
# good_indexed.values => array([set()], dtype=object)

# incorrect
# bad_indexed.values => array([array(5)], dtype=object)
[...]
The current behavior prevents storing objects inside arrays of `dtype==object` even when only performing non-broadcasted assignments if the RHS has a `values` property.
[...]
\end{taskbox}

\paragraph{Folding.}
The agent reads \texttt{as\_compatible\_data} at step~$6$. In its next reply it reports the root cause
and names steps $4$ to $6$ stale, step~$6$ included.

\begin{replybox}{Reply at step 7 of \texttt{pydata\_\_xarray-2905}, the command shortened}
PROGRESS: Found the problematic line [...] in as_compatible_data that extracts .values from any object
STALE: 4, 5, 6
cat > /testbed/test_issue.py << 'EOF'
import xarray as xr
[... 14 lines ...]
EOF
python /testbed/test_issue.py
\end{replybox}

The protocol tells it to keep exactly this read, the one that pinned down the root cause. Steps $1$ to
$5$, named earlier, are already folded, and from step~$12$ on step~$6$ joins them. The six turns,
$13{,}584$ characters in the log, now share one $1{,}219$-character note, and across this fold the prompt
shrinks by more than $2{,}200$ characters, although it carries one more turn.

\begin{notebox}{Steps 1--6 of \texttt{pydata\_\_xarray-2905}, folded from step 12 on}
[context-manager] steps 1-6 folded: Reading variable.py to understand the __setitem__ issue with .values property coercion · Examining the __setitem__ method in variable.py to understand the coercion issue · Now I need to find the as_compatible_data function to understand where the coercion happens. `restore 1-6` shows them again.
  $ ls -la /testbed/xarray/core/ -> total 1236 | drwxrwxrwx 2 root root   4096 Aug 13 04:34 . | drwxrwxrwx 9 root root   4096 Aug 13 04:34 .. | -rw-r--r-- 1 root root      0 Aug 13 04:34 __init__.py
  $ grep -n "__setitem__" /testbed/xarray/co -> 790:        NB. __getitem__ and __setitem__ implement xarray-style ind | 849:    def __setitem__(self, key, value): | 850:        """__setitem__ is overloaded to access the underlying nump | 2553:    def __setitem__(self, key, value):
  variable.py:     def __setitem__(self, key, value):;     def attrs(self) -> Dict[Hashable, Any]:;     def attrs(self, value: Mapping[Hashable, Any]) -> None:;     def encoding(self):;     def encoding(self, value)
  $ grep -n "def as_compatible_data" /testbe -> 189:def as_compatible_data(data, fastpath=False):
  variable.py: def as_compatible_data(data, fastpath=False):; def _as_array_or_item(data):
\end{notebox}

From step~$12$ on, neither the command of step~$6$ nor the line the patch later changes appears anywhere
in the prompt. The note keeps the signature of \texttt{as\_compatible\_data} and, from the grep, its line
number.

\paragraph{Deduplication.}
At step~$12$ the agent rewrites its test script, runs it, and gets a $25$-line traceback. At step~$13$ it
changes the script again and gets the same traceback, character for character. Step~$13$ renders as a note
that points to step~$12$, which still shows the traceback in full, and $1{,}044$ characters shrink to
$130$.

\begin{notebox}{Step 13 of \texttt{pydata\_\_xarray-2905}, a repeat of step 12}
[step 13]
[context-manager] output omitted: identical to step 12. `restore 13` shows it again.
<returncode>1</returncode>
... (25 lines) ...
\end{notebox}

\paragraph{Recovery by re-running.}
At step~$15$ the agent reproduces the bug. At step~$16$ it writes ``Let me look at the function to
understand the fix needed'' and re-issues the step-$6$ command verbatim. The file has not been edited, so
the output is byte-identical to what was folded, and because it repeats folded content it renders in full
rather than as a stub.

\begin{outputbox}{Step 16 of \texttt{pydata\_\_xarray-2905}, the command of step 6 run again}
[step 16]
<returncode>0</returncode>
<output>
def as_compatible_data(data, fastpath=False):
    """Prepare and wrap data to put in a Variable.
[... 29 lines ...]
    # we don't want nested self-described arrays
    data = getattr(data, "values", data)
[... 59 lines ...]
</output>
\end{outputbox}

At steps~$31$ to $34$ the agent patches that line, and the task is solved. The agent's \textsc{stale} was
wrong, and one re-run undid it.

\section{Additional results}
\label{sec:appendix-main}

\subsection{A proprietary model}
\label{sec:appendix-claude}

Table~\ref{tab:claude} runs \strat{base} and ReFold with Claude Opus 4.8, served through Amazon Bedrock
with prompt caching, on $21$ SWE-bench Verified tasks in the same harness at $w{=}8$. ReFold cuts the
tokens of a task by $24\%$, its steps by $17\%$, its cost by $11\%$ and its duration by $19\%$, and resolve stays within the run-to-run noise of Appendix~\ref{sec:appendix-variability}. Claude finishes these tasks in
a median of $18$ steps, against $50$ for Qwen3.6-27B in Table~\ref{tab:main}, and the prompt cache
serves $94\%$ of the prompt tokens of \strat{base} and $92\%$ of those of ReFold, at a tenth of the input
price.

\begin{table}[ht]
  \caption{Claude Opus 4.8 on $21$ SWE-bench Verified tasks at $w{=}8$. ReFold cuts tokens, steps, cost and duration by $11$ to $24\%$.}
  \label{tab:claude}
  \centering \small
  \setlength{\tabcolsep}{5pt}
  \renewcommand{\arraystretch}{1.14}
  \begin{tabular}{lP{1.4pt}P{0pt}P{0pt}P{10.4pt}P{0pt}P{8.1pt}}
    \toprule
    Arm & \multicolumn{2}{c}{Resolve} & \multicolumn{2}{c}{Steps} & \multicolumn{2}{c}{Tokens (k)} & \multicolumn{2}{c}{Completion (k)} & \multicolumn{2}{c}{Cost (\textcent)} & \multicolumn{2}{c}{Duration (min)} \\
    \midrule
    \strat{base} & $13$ & & $18$ & & $273$ & & $4.9$ & & $38.3$ & & $3.2$ & \\
    \rowcolor{black!7}
    ReFold & $12$ & \nd{($-$1)} & $15$ & \gd{($-$17\%)} & $208$ & \gd{($-$24\%)} & $4.5$ & \gd{($-$8\%)}
      & $33.9$ & \gd{($-$11\%)} & $2.6$ & \gd{($-$19\%)} \\
    \bottomrule
  \end{tabular}
\end{table}

\subsection{Per-step growth}
\label{sec:appendix-growth}

\begin{figure}[ht]
  \centering
  \includegraphics[width=\linewidth]{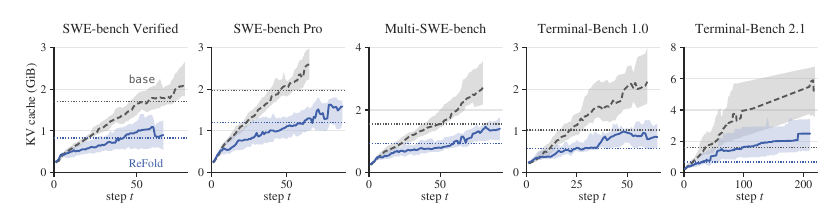}
  \caption{KV cache per session at every step with Qwen3.6-27B. The gap between \strat{base} and ReFold widens with trajectory length, as \strat{base} keeps everything it has read.}
  \label{fig:growth}
\end{figure}

Figure~\ref{fig:growth} traces how the KV cache of a session evolves step by step behind the totals of
Table~\ref{tab:main}, as the median over the tasks still running with the interquartile range shaded
and each arm's median peak dotted. Under \strat{base} it climbs at every step because every step appends its
observation and nothing is removed, so cumulative tokens grow faster than linearly with trajectory
length (Section~\ref{sec:method-setup}). Under ReFold it settles into a slowly rising sawtooth, and the gap
widens with length. On SWE-bench Verified the part of the cache that grows with the prompt is $15\%$
smaller at step $10$, $35\%$ at step $20$ and $46\%$ at step $30$, and every benchmark widens the same way.

\subsection{Per-task distribution}
\label{sec:appendix-dist}

\begin{figure}[ht]
  \centering
  \includegraphics[width=\linewidth]{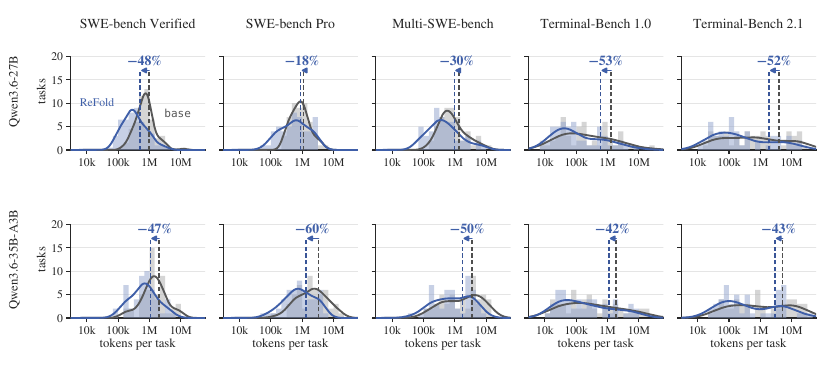}
  \caption{Tokens per task for every cell of Table~\ref{tab:main}. ReFold shifts the whole distribution toward fewer tokens rather than trimming a few long tasks.}
  \label{fig:dist}
\end{figure}

Figure~\ref{fig:dist} shows how the tokens of each cell of Table~\ref{tab:main} are distributed over its
tasks, with each arm's mean dashed and the change of Table~\ref{tab:main} marked between them. ReFold
sends fewer tokens than \strat{base} on $66$ to $90\%$ of the tasks of every cell.

\subsection{Published methods on the same harness}
\label{sec:appendix-external}

Table~\ref{tab:external} compares ReFold with three published methods at the operating point of
Table~\ref{tab:main}, Qwen3.6-27B on SWE-bench Verified at the full window with $w{=}8$. The \strat{base}
and ReFold rows are those of Tables~\ref{tab:main} and~\ref{tab:scarcity}, the mean of their two runs in
Table~\ref{tab:aa}.
LLMLingua-2 \citep{pan2024llmlingua2} compresses the body of every observation of at least $2{,}000$
characters at the token level with its released classifier on a local GPU, keeping half of the tokens,
exempting the three most recent observations and preserving the characters that frame an observation.
SWE-Pruner \citep{wang2026swepruner} is the implementation of Appendix~\ref{sec:appendix-strategies}.
We run AgentDiet \citep{xiao2026agentdiet} with the served model as its reflection model, which at each
step rewrites one earlier observation longer than $\theta$ tokens, given the two steps before it and the
one after, and the rewrite is adopted only if it saves more than $\theta$ tokens. The three most recent observations are exempt and a rewrite is
capped at $2{,}048$ tokens, and we run $\theta{=}500$ and $\theta{=}150$. Its rewriting calls count toward
its tokens and cost and are priced as uncached prompt tokens. In Table~\ref{tab:external}, cost is in US
cents per task and split into the main loop and the method's own model calls, which are zero for the
two methods that run a local model, and ratios are against \strat{base}.

\begin{table}[ht]
  \caption{Published methods at the setting of Table~\ref{tab:main}. ReFold cuts tokens and cost the most, and only ReFold halves the KV cache of a session.}
  \label{tab:external}
  \centering \small
  \setlength{\tabcolsep}{4pt}
  \renewcommand{\arraystretch}{1.14}
  \begin{tabular}{lrP{0pt}rrP{0pt}P{0pt}r}
    \toprule
    & & & & \multicolumn{4}{c}{Cost (\textcent)} & & & \\
    \cmidrule(lr){5-8}
    Method & Resolve & \multicolumn{2}{c}{Tokens (k)} & \multicolumn{1}{c}{main} & \multicolumn{1}{c}{aux}
      & \multicolumn{2}{c}{total} & \multicolumn{2}{c}{Dur.\ (min)} & KV cache (GiB) \\
    \midrule
    \strat{base} & $38$ & $965$ & & $19.5$ & --- & $19.5$ & & $9.8$ & & $1.71$ \\
    LLMLingua-2 & $35$ & $885$ & \gd{($-$8\%)} & $18.6$ & $0$ & $18.6$ & \gd{($0.95\times$)}
      & $10.7$ & \bd{($1.09\times$)} & $1.52$ \\
    SWE-Pruner & $35$ & $905$ & \gd{($-$6\%)} & $18.7$ & $0$ & $18.7$ & \gd{($0.96\times$)}
      & $9.6$ & \nd{($0.98\times$)} & $1.55$ \\
    AgentDiet $\theta{=}500$ & $38$ & $980$ & \nd{($+$2\%)} & $20.6$ & $4.5$ & $25.1$ & \bd{($1.29\times$)}
      & $20.8$ & \bd{($2.12\times$)} & $1.72$ \\
    AgentDiet $\theta{=}150$ & $37$ & $817$ & \gd{($-$15\%)} & $17.5$ & $12.2$ & $29.7$ & \bd{($1.52\times$)}
      & $34.0$ & \bd{($3.46\times$)} & $1.45$ \\
    \rowcolor{black!7}
    ReFold & $38$ & $505$ & \gd{($-$48\%)} & $12.9$ & --- & $12.9$ & \gd{($0.66\times$)}
      & $9.4$ & \gd{($0.96\times$)} & $0.83$ \\
    \bottomrule
  \end{tabular}
\end{table}

ReFold cuts tokens by $48\%$ and cost by $34\%$, several times more than any other method, and only
ReFold halves the memory a session holds, because a method that drops lines or tokens inside an observation cannot remove the whole
repeated observations that set the peak. AgentDiet's rewriting calls are why its cost rises with how
often it is asked, and at $\theta{=}150$ its auxiliary generation is three times that of its main loop.
Resolve stays within $3$ of \strat{base} for every method, inside the run-to-run noise of
Appendix~\ref{sec:appendix-variability}.

\section{Two operators}
\label{sec:appendix-ablation}\label{sec:appendix-operators}

We switch off deduplication or folding in ReFold, with Qwen3.6-27B on SWE-bench Verified at the full
window and $w{=}8$ (Table~\ref{tab:operators}), with changes against the row with both off, which is
\strat{base} of Table~\ref{tab:main}. The row with both on is ReFold of Table~\ref{tab:main}. Folding
does most of the work. With folding alone the KV cache a session holds is $49\%$ below \strat{base} and
the tokens of a task $50\%$, where deduplication alone takes $16\%$ and $33\%$. With both on, the KV
cache falls by $52\%$, the lowest of the four cells. Completion tokens are at most $3\%$ of the tokens in
every cell, so every difference is on the prompt side.

\begin{table}[ht]
  \caption{Deduplication and folding switched off in ReFold one at a time. The row with both off is
    \strat{base} of Table~\ref{tab:main}, and the row with both on is ReFold. Folding does most of the
    reduction, and the two together leave the smallest KV cache.}
  \label{tab:operators}
  \centering \small
  \setlength{\tabcolsep}{6pt}
  \renewcommand{\arraystretch}{1.12}
  \begin{tabular}{ccP{1.4pt}P{8.2pt}P{0pt}}
    \toprule
    Dedup.\ & Fold & \multicolumn{2}{c}{Resolve} & \multicolumn{2}{c}{KV cache (GiB)} & \multicolumn{2}{c}{Tokens (k)} \\
    \midrule
    --- & --- & $38$ & & $1.71$ & & $965$ & \\
    $\checkmark$ & --- & $40$ & \nd{($+$2)} & $1.43$ & \gd{($-$16\%)} & $649$ & \gd{($-$33\%)} \\
    --- & $\checkmark$ & $36$ & \nd{($-$2)} & $0.88$ & \gd{($-$49\%)} & $505$ & \gd{($-$48\%)} \\
    \rowcolor{black!7}
    $\checkmark$ & $\checkmark$ & $38$ & \nd{($\pm$0)} & $0.83$ & \gd{($-$52\%)} & $478$ & \gd{($-$50\%)} \\
    \bottomrule
  \end{tabular}
\end{table}

\section{Composition with other mechanisms}
\label{sec:appendix-compose}

ReFold needs only a rendering hook and prompt instructions, so it can run beneath mechanisms that act
elsewhere in the loop, as it runs beneath compaction in Section~\ref{sec:results-scarcity}. Here it runs beneath three
more, one in the serving system, one in the choice of model and one in the prompt.

\subsection{Routing across replicas}
\label{sec:appendix-replicas}

We serve Qwen3.6-27B from two replicas of two GPUs each and route the calls of every task in two ways
(Table~\ref{tab:replicas}). Sticky routing sends every call of a task to one replica, so its prefix
cache holds the whole trajectory. Round-robin routing alternates the replicas call by call, so a call can
reuse only what its replica cached two calls earlier. At $w{=}16$ round-robin lowers the
cache hit of both arms by a similar amount, $15.4$ points for \strat{base} and $12.5$ for ReFold. At
$w{=}32$ \strat{base} fills the KV pools of both replicas, to $99.7$ and $99.8\%$, and is preempted $10$
and $63$ times, and its cache hit falls from $78.3$ to $13.7\%$. ReFold peaks at $68$ and $80\%$ without
a preemption, keeps a cache hit of $31.2\%$, and resolves $40$ tasks against $26$ in $41.8$ against
$75.5$ minutes. Under sticky routing the two arms resolve
within four tasks of each other at both loads, while ReFold still cuts tokens by $55$ and $42\%$ and duration by $24$ and $34\%$.

\begin{table}[ht]
  \caption{Qwen3.6-27B on two replicas under sticky and round-robin routing. Round-robin lowers the cache hit of both arms, and at $w{=}32$, where \strat{base} fills both KV pools, ReFold resolves $40$ tasks against $26$.}
  \label{tab:replicas}
  \centering \small
  \setlength{\tabcolsep}{5pt}
  \renewcommand{\arraystretch}{1.14}
  \begin{tabular}{llP{0pt}P{0pt}P{4pt}P{5.8pt}}
    \toprule
    Routing, $w$ & Arm & \multicolumn{2}{c}{Resolve} & \multicolumn{2}{c}{Tokens (k)} & \multicolumn{2}{c}{Cache hit (\%)} & \multicolumn{2}{c}{Duration (min)} \\
    \midrule
    Sticky, $w{=}16$ & \strat{base} & $40$ & & $1118$ & & $89.5$ & & $12.5$ & \\
    \rowcolor{black!7} & \quad + ReFold & $36$ & \nd{($-$4)} & $502$ & \gd{($-$55\%)}
      & $76.4$ & \bd{($-$13.2)} & $9.5$ & \gd{($-$24\%)} \\
    \midrule
    Round-robin, $w{=}16$ & \strat{base} & $36$ & & $784$ & & $74.1$ & & $18.2$ & \\
    \rowcolor{black!7} & \quad + ReFold & $36$ & \nd{($\pm$0)} & $683$ & \gd{($-$13\%)}
      & $63.9$ & \bd{($-$10.2)} & $15.6$ & \gd{($-$14\%)} \\
    \midrule
    Sticky, $w{=}32$ & \strat{base} & $38$ & & $898$ & & $78.3$ & & $23.1$ & \\
    \rowcolor{black!7} & \quad + ReFold & $35$ & \nd{($-$3)} & $523$ & \gd{($-$42\%)}
      & $72.2$ & \bd{($-$6.1)} & $15.3$ & \gd{($-$34\%)} \\
    \midrule
    Round-robin, $w{=}32$ & \strat{base} & $26$ & & $623$ & & $13.7$ & & $75.5$ & \\
    \rowcolor{black!7} & \quad + ReFold & $40$ & \gd{($+$14)} & $441$ & \gd{($-$29\%)}
      & $31.2$ & \gd{($+$17.5)} & $41.8$ & \gd{($-$45\%)} \\
    \bottomrule
  \end{tabular}
\end{table}

\subsection{Routing between models}
\label{sec:appendix-models}

One server runs Qwen3.6-27B and a second Qwen3.6-35B-A3B, each on two GPUs. Alternating runs send odd
calls to one model and even calls to the other, so each model sees every other call and reuses only the
prompt it cached two calls earlier (Table~\ref{tab:models}). The \emph{27B only} runs send every call to
Qwen3.6-27B with the second server idle, and repeat
the duration savings of the concurrency rows of Table~\ref{tab:scarcity} at both loads and their resolve
gain at $w{=}32$. Their token saving at $w{=}32$ is smaller, since eight \strat{base} tasks of
Table~\ref{tab:scarcity} end at the step limit against one here. Under the
alternation ReFold cuts tokens
by $59$ and $27\%$ and duration by $26$ and $33\%$ at $w{=}16$ and $32$, and resolves $37$ tasks at both
loads against $33$ and $27$. At $w{=}32$, $13$ alternating and $21$ 27B-only tasks of \strat{base} end
in a timeout, against none and $2$ for ReFold.

\begin{table}[ht]
  \caption{Qwen3.6-27B alone or alternating with Qwen3.6-35B-A3B at every call. ReFold keeps its token and duration savings under the alternation and resolves more tasks than \strat{base} at both loads.}
  \label{tab:models}
  \centering \small
  \setlength{\tabcolsep}{5pt}
  \renewcommand{\arraystretch}{1.14}
  \begin{tabular}{llP{0pt}rP{0pt}P{5.8pt}}
    \toprule
    Models, $w$ & Arm & \multicolumn{2}{c}{Resolve} & Timeouts & \multicolumn{2}{c}{Tokens (k)} & \multicolumn{2}{c}{Duration (min)} \\
    \midrule
    27B only, $w{=}16$ & \strat{base} & $35$ & & $2$ & $1033$ & & $27.0$ & \\
    \rowcolor{black!7} & \quad + ReFold & $36$ & \nd{($+$1)} & $1$ & $469$ & \gd{($-$55\%)} & $15.4$ & \gd{($-$43\%)} \\
    \midrule
    Alternating, $w{=}16$ & \strat{base} & $33$ & & $3$ & $1218$ & & $15.9$ & \\
    \rowcolor{black!7} & \quad + ReFold & $37$ & \nd{($+$4)} & $0$ & $499$ & \gd{($-$59\%)} & $11.8$ & \gd{($-$26\%)} \\
    \midrule
    27B only, $w{=}32$ & \strat{base} & $21$ & & $21$ & $667$ & & $61.6$ & \\
    \rowcolor{black!7} & \quad + ReFold & $33$ & \gd{($+$12)} & $2$ & $585$ & \gd{($-$12\%)} & $37.9$ & \gd{($-$39\%)} \\
    \midrule
    Alternating, $w{=}32$ & \strat{base} & $27$ & & $13$ & $811$ & & $42.7$ & \\
    \rowcolor{black!7} & \quad + ReFold & $37$ & \gd{($+$10)} & $0$ & $592$ & \gd{($-$27\%)} & $28.6$ & \gd{($-$33\%)} \\
    \bottomrule
  \end{tabular}
\end{table}

\subsection{Retrieval memory}
\label{sec:appendix-memory}

The memory layer retrieves chunks of earlier tool outputs by BM25, the top three chunks of $40$ lines
each, and appends them to the end of the rendered prompt, at most $6{,}000$ characters with each chunk
labeled by the command that produced it. We run it with and without ReFold beneath it at a $16$k budget
with $w{=}8$, next to \strat{base} and ReFold at the same budget without compaction
(Table~\ref{tab:memory}). On its own the memory layer resolves $8$ tasks against $4$ and still exhausts
the budget on $39$ of $50$, because what it retrieves adds to the prompt. With ReFold beneath it, resolve rises to $17$, exhaustions fall to $23$ and the median peak KV cache to
$0.98$\,GiB. ReFold alone does better still, with $21$ resolved and $14$ exhausted, since the retrieved
chunks refill the budget that ReFold frees.

\begin{table}[ht]
  \caption{Retrieval memory and ReFold switched independently at a $16$k budget. ReFold beneath the memory layer doubles its resolve, from $8$ to $17$, and cuts its exhaustions from $39$ to $23$.}
  \label{tab:memory}
  \centering \small
  \setlength{\tabcolsep}{5pt}
  \renewcommand{\arraystretch}{1.12}
  \begin{tabular}{ccrrrrrr}
    \toprule
    Memory & ReFold & Resolve & Exhausted & Tokens (k) & KV cache (GiB) & Steps & Duration (min) \\
    \midrule
    --- & --- & $4$ & $43$ & $288$ & $1.12$ & $31$ & $5.2$ \\
    $\checkmark$ & --- & $8$ & $39$ & $307$ & $1.12$ & $29$ & $14.6$ \\
    --- & $\checkmark$ & $21$ & $14$ & $334$ & $0.87$ & $32$ & $6.9$ \\
    $\checkmark$ & $\checkmark$ & $17$ & $23$ & $616$ & $0.98$ & $52$ & $21.8$ \\
    \bottomrule
  \end{tabular}
\end{table}

\end{document}